\documentclass[letterpaper, 10pt, journal, twoside]{style/IEEEtran}    

\makeatletter
\let\NAT@parse\undefined
\makeatother

\usepackage{dblfloatfix}

\usepackage[numbers,sectionbib,sort&compress]{natbib}
\usepackage{bm}
\usepackage{pgfplots}
\usepackage{gensymb}
\usepackage{svg}
\usepackage{graphicx}
\usepackage{amsmath}
\usepackage{amssymb}
\usepackage{algorithm}
\usepackage[noend]{algpseudocode}
\usepackage{subcaption}
\usepackage{amsfonts}
\usepackage{siunitx}
\usepackage{booktabs}
\usepackage{makecell}
\usepackage{multirow}
\usepackage{upgreek}
\usepackage[font=small]{caption}
\usepackage[export]{adjustbox}
\usepackage{tikz}
\usepackage{sidecap} \sidecaptionvpos{figure}{c}
\newcommand*{\dd}[2][]{\mathop{\mathrm{d}^{#1}#2}}

\newcommand*{\x}[1][]{\bm{x}_{#1}}
\newcommand*{\GP}{\mathcal{GP}}
\newcommand*{\Real}{\mathbb{R}}
\newcommand*{\Env}{\mathcal{E}}
\newcommand*{\cell}[1][]{\mathit{c}_{#1}}
\newcommand*{\map}{\mathit{C}}

\newcommand*{\area}[1][]{A_{#1}}
\newcommand*{\prior}[1]{{#1}^-}
\newcommand*{\posterior}[1]{{#1}^+}

\usepackage{hyperref}
\hypersetup{colorlinks,breaklinks,
linkcolor=[rgb]{0.5,0.,0.},
citecolor=[rgb]{0.000,0.427,0.173},
urlcolor=[rgb]{0.031,0.318,0.612}}

\usepackage[nameinlink, capitalize]{cleveref}
\usepackage{color}
\definecolor{CommentPink}{rgb}{1,0.2,0.5}
\definecolor{CommentBlue}{rgb}{0,0,1}
\definecolor{CommentGreen}{rgb}{0,1,0}

\Crefname{section}{Sec.}{Secs.}
\Crefname{table}{Tab.}{Tabs.}

\usepackage[printonlyused,withpage,nolist,nohyperlinks]{acronym}

\IEEEoverridecommandlockouts                     

\title{\LARGE \bf
Adaptive-Resolution Field Mapping \\ Using Gaussian Process Fusion with Integral Kernels}

\author{Liren Jin$^{1}$, Julius R\"{u}ckin$^{1}$, Stefan H.\ Kiss$^{2}$, Teresa Vidal-Calleja$^{2}$, Marija Popovi\'{c}$^{1}$
\thanks{$^{1}$Cluster of Excellence PhenoRob, Institute of Geodesy and Geoinformation, University of Bonn. $^{2}$UTS Robotics Institute, Faculty of Engineering and IT, University of Technology Sydney. This work was funded by the Deutsche Forschungsgemeinschaft (DFG, German Research Foundation) under Germany’s Excellence Strategy - EXC 2070 – 390732324. Corresponding: \texttt{ljin@uni-bonn.de}.}
}

\begin{document}

\maketitle

\begin{abstract}
Unmanned aerial vehicles are rapidly gaining popularity in a variety of environmental monitoring tasks.
A key requirement for their autonomous operation is the ability to perform efficient environmental mapping online, given limited on-board resources constraining operation time, travel distance, and computational capacity.
To address this, we present an online adaptive-resolution approach for mapping terrain based on Gaussian Process fusion. 
A key aspect of our approach is an integral kernel encoding spatial correlation over the areas of grid cells, which enables modifying map resolution while maintaining correlations in a theoretically sound fashion.
This way, we can retain details in areas of interest at higher map resolutions while compressing information in uninteresting areas at coarser resolutions to achieve a compact map representation of the environment.  
We evaluate the performance of our approach on both synthetic and real-world data. Results show that our method is more efficient in terms of mapping time and memory consumption without compromising on map quality. Finally, we integrate our mapping strategy into an adaptive path planning framework to show that it facilitates information gathering efficiency in online settings.  

\end{abstract}

\section{Introduction} \label{S:introduction}

Environmental monitoring plays a central role in helping us better understand the Earth and its natural processes. However, many commonly observed natural phenomena, e.g., temperature, humidity, etc., exhibit complex spatial variations that are difficult to capture using traditional sensing methods, such as manual sampling or static sensor networks \citep{Dunbabin2012,Manfreda2018,Tmusic2020}. Recently, unmanned aerial vehicles (UAVs) have emerged as a more flexible, cost-efficient alternative for data acquisition in a wide range of applications, including biomass measurement~\citep{Popovic2017a,Manfreda2018}, signal strength monitoring~\citep{Hollinger2014}, weed detection~\citep{Stache2021}
and thermal mapping~\citep{Bircher2016,Manfreda2018}. To fully exploit these platforms, a key challenge is developing map representations that can accurately capture heterogeneous natural variables, while also being compact and computationally efficient for online interpretability and decision-making on resource-constrained systems.

\begin{figure}[!t]
\centering
  \begin{subfigure}[]{0.24\textwidth}
  \includegraphics[width=\textwidth]{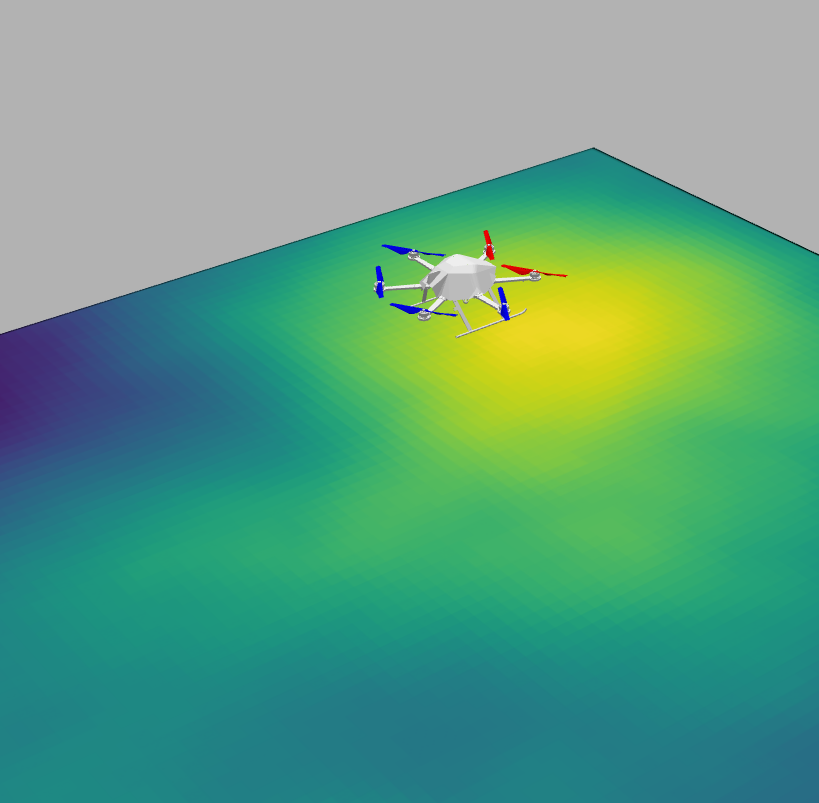}
  \end{subfigure}\hfill%
  \begin{subfigure}[]{0.235\textwidth}
  \includegraphics[width=\textwidth]{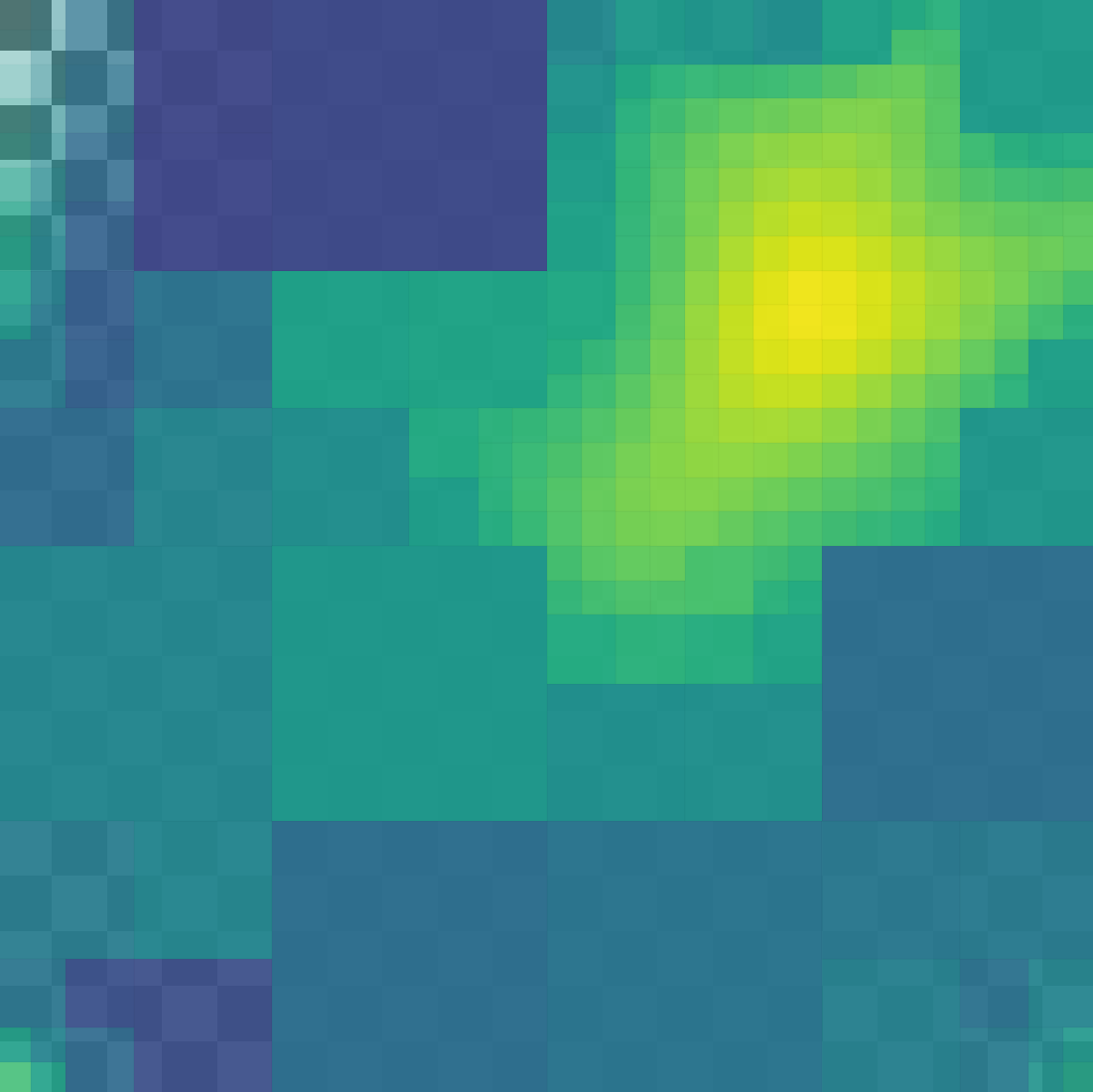}
  \end{subfigure}\hfill%
    \caption{Our adaptive-resolution Gaussian Process fusion approach for online field mapping. Left: Synthetic ground truth distribution. Yellower shades indicate higher values we would like to map in greater detail. Right: Mapping result with uncertainty. Our approach maps areas of interest at higher resolutions while compressing information in less interesting regions to increase computational and memory efficiency. The checkerboard is added for visual interpretation of the map uncertainty (high opacity means low uncertainty).}\label{F:teaser}
\end{figure}
This paper focuses on mapping methods for terrain monitoring scenarios, where the aim is to recover a continuous, non-uniform 2D scalar field, e.g., of temperature, biomass cover, etc., using measurement information from on-board sensors. In this setup, our goal is to develop a mapping strategy that can accommodate \textit{both} high-fidelity field reconstruction in targeted areas of interest, e.g., hotspots or anomalies, as well as mapping with low computational and memory requirements. By catering for these two aspects simultaneously, our work bridges the gap between environmental monitoring problems and autonomous robotic applications, e.g., adaptive path planning based on the current map state.

There are several methods for field mapping in environmental monitoring contexts. In the remote sensing community, most existing approaches exploit aerial data to create high-resolution reconstructions, e.g., terrain orthomosaics~\citep{Tmusic2020,Manfreda2018}. Although they produce very detailed models, such procedures often involve heavy postprocessing and are thus not suitable for online applications. A common strategy to tackle this problem is to discretize the environment in a grid map and fuse new measurements into it during a monitoring mission. However, traditional grid-based methods \citep{Hornung2013,Funk2021,Einhorn2011} assume independence between cells, neglecting important spatial correlations which characterize environmental phenomena, and thereby limiting the map quality. 

We propose a new method for mapping continuous fields online.
Our approach is based on \textit{Gaussian Process (GP) fusion} \citep{Teresa2014,Popovic2020}: we exploit a GP model to capture the spatial correlations in the underlying field and use it as a prior for recursive Bayesian fusion.
In this setting, our key goal is to \textit{adaptively} adjust the map resolution online based on the information value of associated measurements, such that only areas of interest are mapped at higher resolutions.
Different from traditional GP regression, which pools the entire measurement history to predict the posterior map state at any resolution at once, the usage of GP fusion, 
although more efficient, poses a major challenge: in order to account for resolution changes, we need to modify online not only the map mean, but also the covariance while properly updating correlations between cells. In other words, adapting the map resolution leads to varying map query points in the environment; 
however, spatial correlations at these new points cannot be easily obtained from the previous measurements or the current map state~\citep{Reece2010}.
The covariance of an adaptive-resolution map is thus difficult to retrieve in a theoretically sound and efficient manner and constitutes an open research question.

To address this, we propose a novel approach based on an \textit{integral kernel} describing the spatial correlation over the areas of grid cells instead of query points, e.g., grid cell center point. Combined with a ND-tree structure, we can adapt the map resolution online while preserving its spatial correlations.
This enables us to retain high-resolution details in targeted areas of the field, while using coarser resolutions otherwise, as shown in \cref{F:teaser}. This way, we achieve memory and computationally efficient mapping without sacrificing on map quality, as necessary for online application on platforms, e.g., UAVs, with limited computing power.
In sum, our contributions are:
\begin{enumerate}
    \item A new method for incrementally mapping continuous scalar fields online. Our approach combines GP fusion with the ND-tree data structure to allow for efficiently changing map resolution based on incoming sensor data. 
    \item An integral kernel function to encode the spatial correlation of 2D grid cells in a continuous field map. This enables us to merge grid cells at any scale to compress information in uninteresting regions, while preserving spatial correlations in the map.
\end{enumerate}
Our mapping approach is evaluated and benchmarked against state-of-the-art approaches using synthetic and real-world data. Experimental results show that our method reduces memory consumption and improves computational efficiency when compared against mapping baselines. Further, we demonstrate its applicability for online adaptive path planning.

\section{Related Work} \label{S:related_work}
A large body of literature has studied mapping methods for monitoring continuous phenomena in different application domains \cite{Stachniss2009,Gregory2014,Hitz2017,Popovic2017a,Shrihari2009,Teresa2014}. Our work focuses on online mapping methods suitable for robotic monitoring scenarios. Our new approach introduces an integral kernel for adaptive-resolution mapping which brings together two key concepts: (1) GP models and (2) ND-tree structure. The following subsections review previous studies related to these topics.

\subsection{Gaussian Processes Mapping}
Grid maps are the most commonly used representation for robotic mapping~\citep{Elfes1989}. Despite their successful application, traditional occupancy grid models assume the stochastic independence of grid cells to enhance computational efficiency~\citep{Simon2012}. However, this representation often poorly captures the spatial correlations found in natural physical phenomena, e.g., distributions of temperature, humidity, etc. To address this, GP models are applied in environmental monitoring. For instance, GPs are used to incorporate uncertainty and represent spatially-correlated data in 2.5D pipe thickness mapping~\citep{Teresa2014}. \citet{Shrihari2009} apply GP regression to predict elevation on a field where sensory information is incomplete. Other applications include gas distribution mapping~\citep{Stachniss2009}, occupancy mapping~\citep{Simon2012} and aquatic monitoring~\citep{Gregory2014,Hitz2017}. Our work follows these lines by using GPs to model a scalar field. 

The main limitation of applying standard GP regression for online robotic mapping is its cubically growing computational complexity as measurements accumulate over time \citep{Rasmussen2006}.
Previous work has tackled this problem by storing measurements in a KD-tree structure
and using local models to approximate GPs \citep{Simon2012, Shen2005, Shrihari2009}. To predict the mean and variance of query points, only nearby measurements are considered. However, local GPs require performing regression for each query point individually. To alleviate this problem, the concept of extended blocks was introduced \citep{Kim2014}, which applies GPs to the query points in individual blocks of the map only using the measurements in neighboring blocks. This approach decomposes a large GP into sub-models and applies regression to infer the posterior of each block. The multiple regression results are then fused using a Bayesian Committee Machine (BCM)~\citep{VOlker2000}, whose computational complexity scales cubically with the number of query points. Based on that, \citet{Wang2016} introduce test-data octrees, which prune nodes of the same state to condense the number of query points in regression.

In GP-based occupancy grid mapping with range sensors, \citet{OCallaghan2011} propose an integral kernel to handle beam line observations directly rather than discretizing them into point observations, thereby reducing the number of measurements used for GP regression.
Most similar to our approach, \citet{Reid2013} use an integral kernel to capture spatial correlations between image areas and infer a high-resolution estimate from a low-resolution observations in a UAV-based setup. However, inference over the map is still performed using standard GP regression, which suffers from poor scalability, especially with dense image data.

In contrast to regression-based methods, our method leverages GP fusion~\citep{Popovic2020,Teresa2014} to reduce the computational burden for online mapping. Namely, we exploit a GP as a prior and apply recursive filtering to fuse new measurements incrementally into the map. This procedure removes the need to preserve the measurement history and infer the map posterior distribution from scratch each time new data arrive~\citep{Reece2010}. 
A key difference in our approach with respect to previous fusion-based works~\citep{Popovic2020,Teresa2014} is the proposed integral kernel, which bridges the gap between GP fusion and online adaptive-resolution mapping.

\subsection{Multi-Resolution Mapping}
In practice, many monitoring scenarios exhibit a non-uniform distribution of information in the environment, i.e., some regions are considered more interesting or informative for mapping than others. Therefore, maintaining a map with constant resolution over the whole environment is redundant and costly. A common method to generate compact map representations is by using tree structures. A well-known algorithm in this category is OctoMap~\citep{Hornung2013}, which prunes child nodes with the same state, e.g., occupied, to achieve both memory-savings and highly precise maps. \citet{Funk2021} use the octree structure in an online mapping system that adjusts map resolution based on occupancy state.
Similarly, \citet{Chen2015} apply quadtrees to build multi-resolution 2D maps. The ND-tree generalizes these approaches by subdividing any $d$-dimensional volume recursively with $N^{d}$ children~\citep{Einhorn2011}. Rather than compressing a map only in a postprocessing step, as in OctoMap, we adapt the map resolution online based on incoming measurements similarly to~\citet{Einhorn2011} and \citet{Funk2021}. Our approach shares the same motivation, as we tailor the map structure to conserve memory and computation time in applications requiring online mapping, such as adaptive path planning~\citep{Popovic2020, Hitz2017, Hollinger2014}. 


Previous works in adaptive-resolution mapping assume cell independence \citep{Funk2021,Hornung2013,Einhorn2011}, such that no correlation information needs to be maintained. This substantially simplifies mapping at the cost of map quality. However, in our online mapping setup, the covariance must be modified to account for resolution changes, which is challenging in the GP fusion framework. In a similar setting, \citet{Popovic2017a} introduce an approach for incrementally fusing variable-resolution measurements into a spatially-correlated map. However, their method still considers a fixed-resolution map. In contrast, our new strategy supports adaptive-resolution mapping while preserving spatial correlations, thereby reducing memory usage and improving computational efficiency.

\section{Adaptive-Resolution Gaussian Process Fusion} \label{S:approach}
This section introduces our online field mapping approach based on GP fusion. We define the map using a GP prior and store it in an ND-tree structure. This map is then recursively updated with new measurements using Bayesian fusion.
We first present the theory behind GPs with the integral kernel function and define an `average measurement' model, in which the state of a grid cell represents the average value of a latent scalar function in the area it covers. Then, we explain our Bayesian fusion update and the merging operation for incrementally building multi-resolution field maps. Bringing together these elements, a key contribution in our approach is the ability to efficiently merge grid cells without losing spatial correlations. Note that our setting in this work considers a UAV-based terrain mapping scenario. However, our approach is also applicable for general 2.5D mapping problems.

\subsection{Gaussian Processes and Integral Kernel} \label{SS:gaussian_processes_and_integral_kernel}
A GP is the generalization of a Gaussian distribution over a finite vector space to an infinite-dimensional function space. It is fully described by its mean $\mu(\x)$ and covariance function $k(\x,\,\x')$. In practice, a GP is used to model spatial correlations in a probabilistic non-parametric manner and infer function values at a finite set of query points given observed data~\citep{Rasmussen2006}. The mapping target in our problem is assumed to be a continuous function described by a GP: $f(\x) \sim \GP(\mu, k): \Env \rightarrow \Real$, where $\Env \subset \Real^2$ is the 2D rectangular input space and $\x \in \Env$. 

Given a pre-trained kernel function $k(\x,\,\x')$, we could obtain the correlations between the function values at any two points in our input space. However, a kernel function defined on points limits the ability to handle resolution change in our GP fusion setting with incremental updates. Previous studies in GP fusion \citep{Popovic2020,Teresa2014} exploit a GP as the prior in predefined query positions, e.g., the centers of grid cells. The posteriors at these query points are then recursively updated with grid cell measurements using Bayesian fusion. When adapting the resolution online, we need the map posterior to be available in new query positions, so that recursive update with new measurements can be performed. However, this cannot be achieved efficiently under current setting. 

To address this problem, we propose a new GP fusion approach leveraging an integral kernel. We first sequentially discretize the input space into grid cells using an ND-tree until maximal depth $t$ is reached. Note that only the leaf grid cells are shown and updated in the map $\map = \{\cell[1],\,\dots,\,\cell[n]\}$, where $n = (N^{d})^t$ with $d = 2$ as we focus on 2D field mapping; $\cell[i] = [x_i^\text{min},\, x_i^\text{max}] \times [y_i^\text{min},\, y_i^\text{max}]$ is the parametrization of grid cell $\cell[i] \subset \Env$. We also define $\bm{C} = [\cell[1],\,\dots,\,\cell[n]]^\top$ as the vectorization of $\map$.
Similar to~\citet{Reid2013}, we then modify the learned kernel function $k(\x,\,\x')$ to correctly encode the spatial correlations between areas of grid cells in our mapping framework.
We define ${\zeta(R)} = \frac{1}{\area}\int_{R} f(\x) \dd{\x}$ to represent the average of the latent function $f$ over a rectangular domain $R \subset \Real^2$ with area $\area \in \Real$. Since applying a linear operator to a GP leads to another GP~\citep{Sarkka2011}, we obtain the new GP $\zeta(R) \sim \GP(\mu_{\text{I}}, k_{\text{I}})$, whose mean and covariance function are described as follows:
\begin{align}
    \mu_{\text{I}}(R_i) &= \frac{1}{\area[i]}\int\limits_{R_i} \mu(\x)\dd{\x} \label{E:average_mean} \,,\\
    k_{\text{I}}(R_i, R_j) &= \frac{1}{\area[i] \area[j]}\iint\limits_{R_i \times R_j} k(\x, \,\x') \dd{\x}\dd{\x'} \,,\label{E:average_kernel}
\end{align}
%
where $\x$ and $\x'$ are the point positions contained within the rectangular domains $R_i$ with area $A_i$ and $R_j$ with area $A_j$ respectively.
The area-related terms in \cref{E:average_mean,E:average_kernel} simply transform the integral into average, which makes the physical meaning of our mean and covariance in accordance with our measurement model introduced in \cref{SS:sensor_model}. This way, we encode the spatial correlations over the area of grid cells in our map, which enables merging operation presented in \cref{SS:merging}

For a grid map with rectangular cells and spatial correlations defined by a squared exponential (SE) kernel, we can find a closed-form solution to \cref{E:average_kernel}. In general, numerical integration is required to determine the kernel integration \citep{OCallaghan2011}. Note that, in our fusion framework, the integral calculation only initializes the map prior and does not burden online mapping.


\subsection{Sensor Model} \label{SS:sensor_model}
\begin{figure}
\centering
   \includegraphics[width=0.35\textwidth, height=4cm]{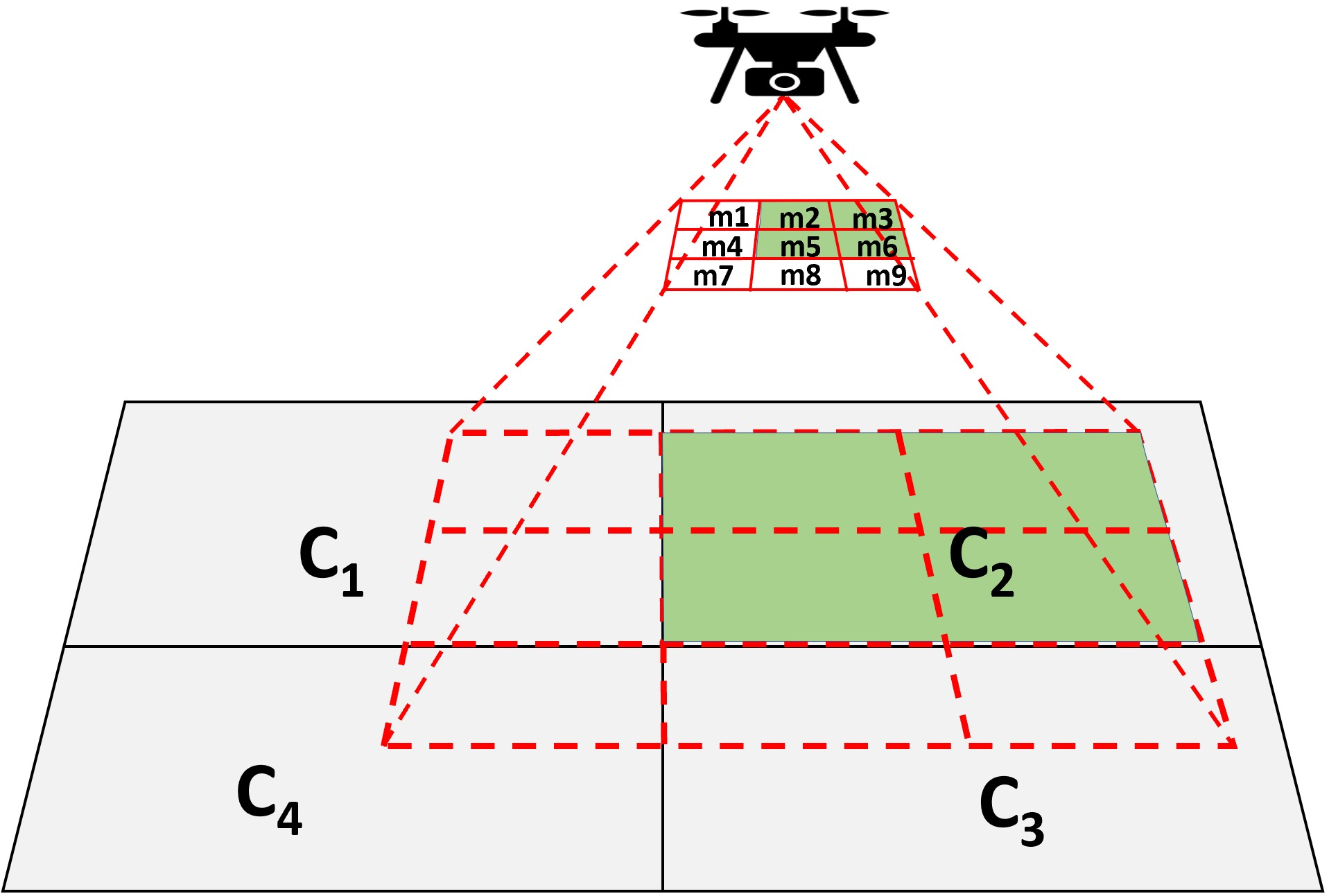}
   \caption{The `average' sensor model provides the measurements of averaged function value over a grid cell. For instance, the measurement $z_2$ observed from $\cell[2]$ is the average of 4 single measurement values $\{m_2,\, m_3,\, m_5,\, m_6\}$. For calculating $\sigma_{c,i}^2$, $\area[\mathrm{cover}]$ is the green area on the terrain and $\area[\cell]$ is the area of $\cell[2]$ itself.
   }\label{F:sensor_model}
\end{figure}
In our GP fusion setting, we consider a Gaussian sensor model to account for noisy measurement data. For each observed grid cell $\cell[i] \in \map$, the sensor provides a measurement $z_i$ capturing the average value of function $f$ over the area of this cell as
$z_i \sim \mathcal{N}(\mu_{s,i}, \sigma_{s,i}^2)$, where $\mu_{s,i}$ is the mean and $\sigma_{s,i}^2$ is the noise variance expressing uncertainty in $z_i$. 

The noise variance can be decomposed into two parts. First, we assume measurements taken from higher altitudes are more susceptible to environmental noise such as light conditions. 
To this end, we describe the degraded accuracy of sensor information at higher altitude 
by $\sigma_{a,i}^2 = \alpha h$,
%
%
where $\alpha \in \mathbb{R}^{+}$ and $h$ is the sensor's altitude. 
Second, we consider uncertainty caused by observing incomplete grid cells. In our mapping framework,
some grid cells are only partially covered by the current sensor footprint, especially when the grid cells occupy larger area after they are merged. Directly assigning the average measurements as the observation of these grid cells would be an over-confident assumption, as the unobserved part of these grid cells may contradict the current measurements, e.g., when grid cells span over the domain of heterogeneous function values. To tackle this problem, we propose the coverage-ratio-dependent variance $\sigma_{c,i}^2 = \beta\left(1 - \frac{\area[\mathrm{cover}]}{\area[c]}\right)$ in our sensor model, 
%
%
where $\beta \in \Real^{+}$ is a weighting and $\area[\cell], \area[cover]$ are the area of the grid cell and the part covered by the footprint. When a grid cell is fully observed in measurement $z_i$, this noise term disappears.

The measurement data in our sensor model are generated as follows: the sensor footprint is determined given the known intrinsic and extrinsic parameters. First, we query the grid cells having overlap with the footprint using depth-first tree search with pruning. Second, for observed grid cell $\cell[i]$, we calculate the corresponding averaged measurement value $z_{i}$ as illustrated in \cref{F:sensor_model}. Third, we sum the altitude-dependent variance $\sigma_{a,i}^2$ and coverage-ratio-dependent variance $\sigma_{c,i}^2$ as the total variance of each measurement $z_i$.

\subsection{Sequential Data Fusion} \label{SS:sequential_data_fusion}
A major difference between GP regression and our GP fusion approach lies in the map update rule. 
In our framework, the map state is fully described by the mean vector $\bm{\mu}(\bm{C})$ and covariance matrix $\bm{K}(\bm{C}, \bm{C})$. Our initial map state can be obtained from \cref{E:average_mean,E:average_kernel}, which is then used as a prior for sequentially fusing new measurements:
\begin{align}
    \prior{\bm{\mu}} = \begin{bmatrix}  
           \mu_{\text{I}}(\cell[1])\\
           \vdots \\
           \mu_{\text{I}}(\cell[n])
        \end{bmatrix}\, ,
    \quad
    \prior{\bm{K}} = \begin{bmatrix}
        k_{\text{I}}(\cell[1], \cell[1])  & \dots  & k_{\text{I}}(\cell[1], \cell[n]) \\
        \vdots & \ddots & \vdots \\
        k_{\text{I}}(\cell[n], \cell[1])  & \dots  &  k_{\text{I}}(\cell[n], \cell[n])
    \end{bmatrix}\, .
\end{align}
%
 We define $\bm{z}$ to be a vector consisting of $m$ new average function value measurements observed from $m$ corresponding grid cells as introduced above. To compute the posterior density $p(\bm{\zeta}|\bm{z},\bm{C}) \propto p(\bm{z}|\bm{\zeta},\bm{C}) \cdot p(\bm{\zeta}|\bm{C})$, we directly apply the Kalman Filter update equations~\citep{Reece2010}:
\begin{align}
 \posterior{\bm{\mu}} ={}& \prior{\bm{\mu}} + \bm{\Gamma}\bm{v} , \label{E:kf_mean} \\
  \posterior{\bm{K}} ={}& \prior{\bm{K}} - \bm{\Gamma}\bm{H}\prior{\bm{K}} \,,\label{E:kf_cov}
\end{align}
%
%
where $\bm{\Gamma} = \prior{\bm{K}}\bm{H}^\top \bm{S}^{-1}$ is the Kalman gain; $\bm{v} = \bm{z} - \bm{H}\prior{\bm{\mu}}$ and $\bm{S} = \bm{H}\prior{\bm{K}}\bm{H}^\top + \bm{R}$ are the measurement and covariance innovations;
$\bm{R}$ is a diagonal $m \times m$ matrix composed of variance term $\sigma^2_{a,i} + \sigma^2_{c,i}$ associated with each measurement $z_i$ and $\bm{H}$ is a $m\times n$ observation matrix denoting the part of the map observed by $\bm{z}$, where $n$ and $m$ are the number of grid cells in the current map and observed grid cells respectively. Note that the current map only contains leaf grid cells and a small matrix $\bm{S} \in \mathbb{R}^{m \times m}$ is inverted at each update.

\subsection{Merging} \label{SS:merging}
A key requirement for multi-resolution mapping is the ability to manipulate the grid cell sizes on-the-fly. Given a non-uniform target field for mapping, our goal is to use coarser (larger) grid cells to map uninteresting regions and denser (smaller) grid cells to retain details in interesting parts. Previous works in GP fusion \citep{Popovic2020,Teresa2014} do not support efficient resolution changes, as the spatial correlations are only maintained in predefined grid cell center points. By using the new GP fusion with integral kernel, however, we naturally encode the states of parent nodes in their children, which enables efficiently retrieving the parent's posterior covariance and mean value from their children on-the-fly. 

\begin{figure}[h]
\centering
   \includegraphics[width=7cm, height=3.5cm]{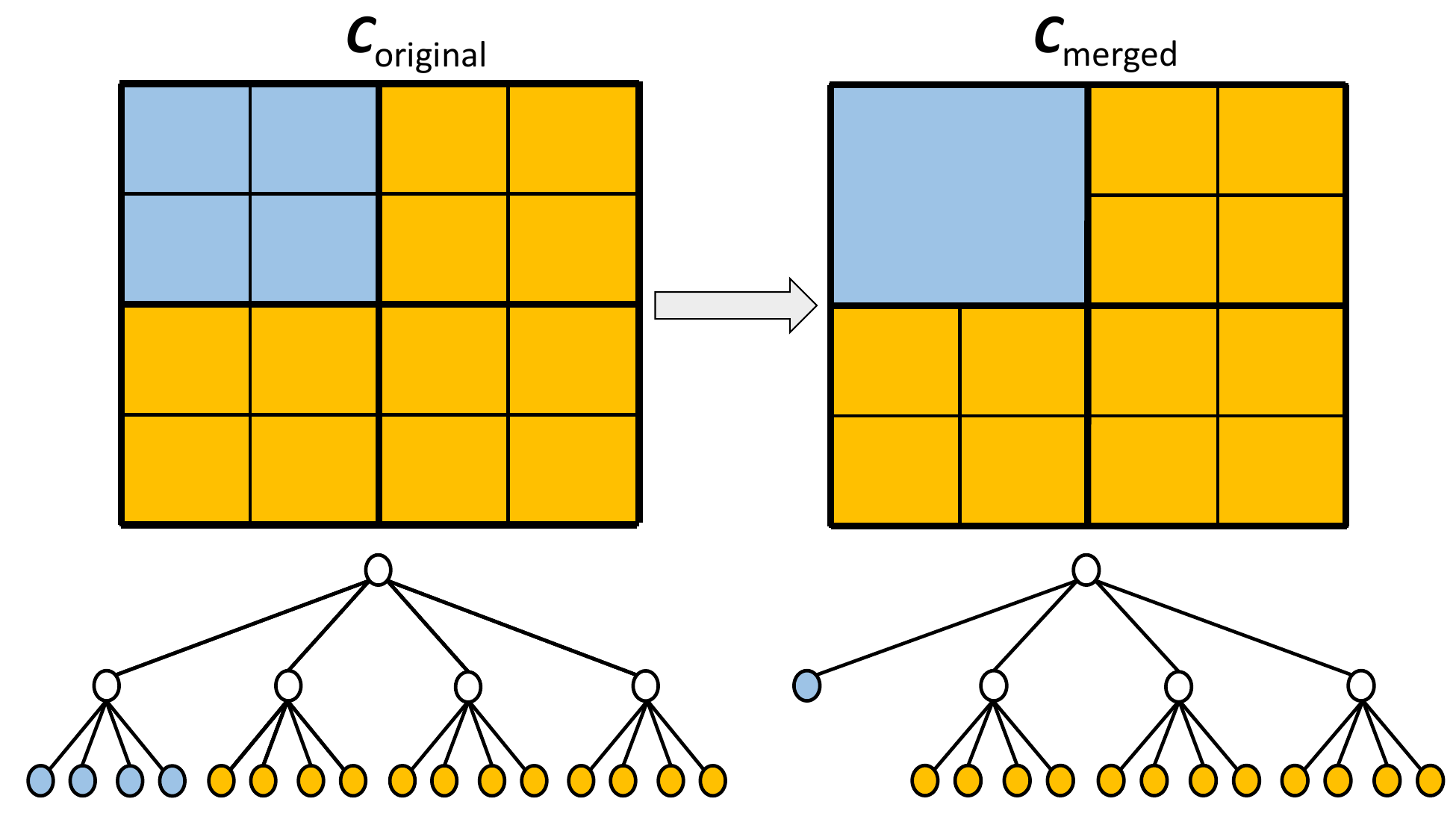}
   \caption{Illustration of our procedure for merging cells in a spatially-correlated map. Top and bottom rows show the grid cell map and its corresponding ND-tree (with $N = d =2$) structure. Only leaf nodes in the tree are shown in the map and updated during Bayesian fusion. After merging (right), the children grid cells are replaced by their parent in the new map.
   }\label{F:tree}
\end{figure}

The merging operation allows us to summarize information in larger areas and thus monotonically reduce the total number of grid cells in the map, which facilitates the mapping efficiency and memory usage. For this, we subdivide our map into uninteresting regions (UR) and hotspots (HS):
\begin{equation} \label{E: hotspots}
    \map_{UR} = \{\cell[i] \in \map ~ \vert ~ \posterior{\bm{\mu}_{i}} + \gamma \posterior{\bm{K}_{i, i}} \leq f_{th}\}, \map_{HS} = \map \setminus \map_{UR} ,
\end{equation}
where $\posterior{\bm{\mu}_{i}}$ and $\posterior{\bm{K}_{i, i}}$ are the posterior mean and variance of grid cell $\cell[i]$; the design parameter $\gamma$ is chosen to specify the margin to the threshold $f_\text{th}$~\citep{Gregory2014}. This setting avoids merging grid cells with possibly high mean values, which would cause detail loss in interesting regions. Thus, grid cells are only merged if there is a strong belief that they fall within uninteresting bounds.
%

For a parent grid cell, if all of its $P = N^d$ child grid cells are uninteresting leaves (grid cells in $\map_{UR}$),
%
%
%
these child grid cells can be replaced by their parent grid cell.
When we merge the information of $P$ children into their parent, based on the definition of grid cell variable and the correlation encoded by the integral kernel, we have the parent grid cell defined as:
\begin{align}
    \zeta_\text{parent} = \frac{1}{P} \sum_{i=1}^{P} \zeta_{\text{child}_i} .
\end{align}
The parent grid cell now represents the average function value of the entire region covered by its children. 
For the grid map, the merging operation can be described as a linear transformation of the GP as follows:
\begin{align}
    \posterior{\bm{\mu}}(\bm{C}_\text{merged}) &=  \bm{M}\posterior{\bm{\mu}}(\bm{C}_\text{original}) \label{E:mean_merge}\,,\\
    \posterior{\bm{K}}(\bm{C}_\text{merged}, \bm{C}_\text{merged}) &= \bm{M}\posterior{\bm{K}}(\bm{C}_\text{original}, \bm{C}_\text{original})\bm{M}^\top . \label{E:covariance_merge}
\end{align}
where $\bm{C}_\text{original}$ represents the vectorized map including $n$ grid cells before merging and $\bm{C}_\text{merged}$ is the newly-merged vectorized map. 
In the simplest case, where only one parent's children cells are merged, $\bm{C}_\text{merged}$, $\bm{C}_\text{original}$, and $\bm{M}$ can be expressed as:
\begin{align}
    \bm{C}_\text{original} &= \begin{bmatrix}
        \cell[1] \\
        \vdots \\
        \cell[n-P]\\
        \cell[n-P+1]\\
        \vdots\\
        \cell[n]
    \end{bmatrix} ,
    &
    \bm{C}_\text{merged} &= \begin{bmatrix}
        \cell[1]\\
        \vdots\\
        \cell[n-P]\\
        \cell[n-P+1]
    \end{bmatrix} ,
\end{align}
\begin{align}
    \bm{M} = \begin{bmatrix}
        \underset{(n-P) \times (n-P)}{\bm{I}} & \underset{(n-P) \times P}{\bm{0}}\\
        \underset{1 \times (n-P)}{\bm{0}} & \underset{1 \times P}{\bm{Q}}
    \end{bmatrix} ,
\end{align}
assuming that $\cell[n-P+1]$ in $\bm{C}_\text{merged}$ represents the parent of grid cells $\{\cell[n-P+1] ,\dots ,\cell[n]\}$ in $\bm{C}_\text{original}$ and $\bm{Q}$ is $\left[\frac{1}{P} ,\, \dots ,\,\frac{1}{P}\right]$. A simple illustration is given in \cref{F:tree}.
The merging operation is performed after every measurement update for eligible grid cells. As the linear transformation of GP leads to another GP, we now treat the resulting GP after the merging operation as a prior map for the next Bayesian update cycle. 

\begin{figure*}[!t]
\centering
\begin{subfigure}{.44\columnwidth}
\includegraphics[width=\columnwidth]{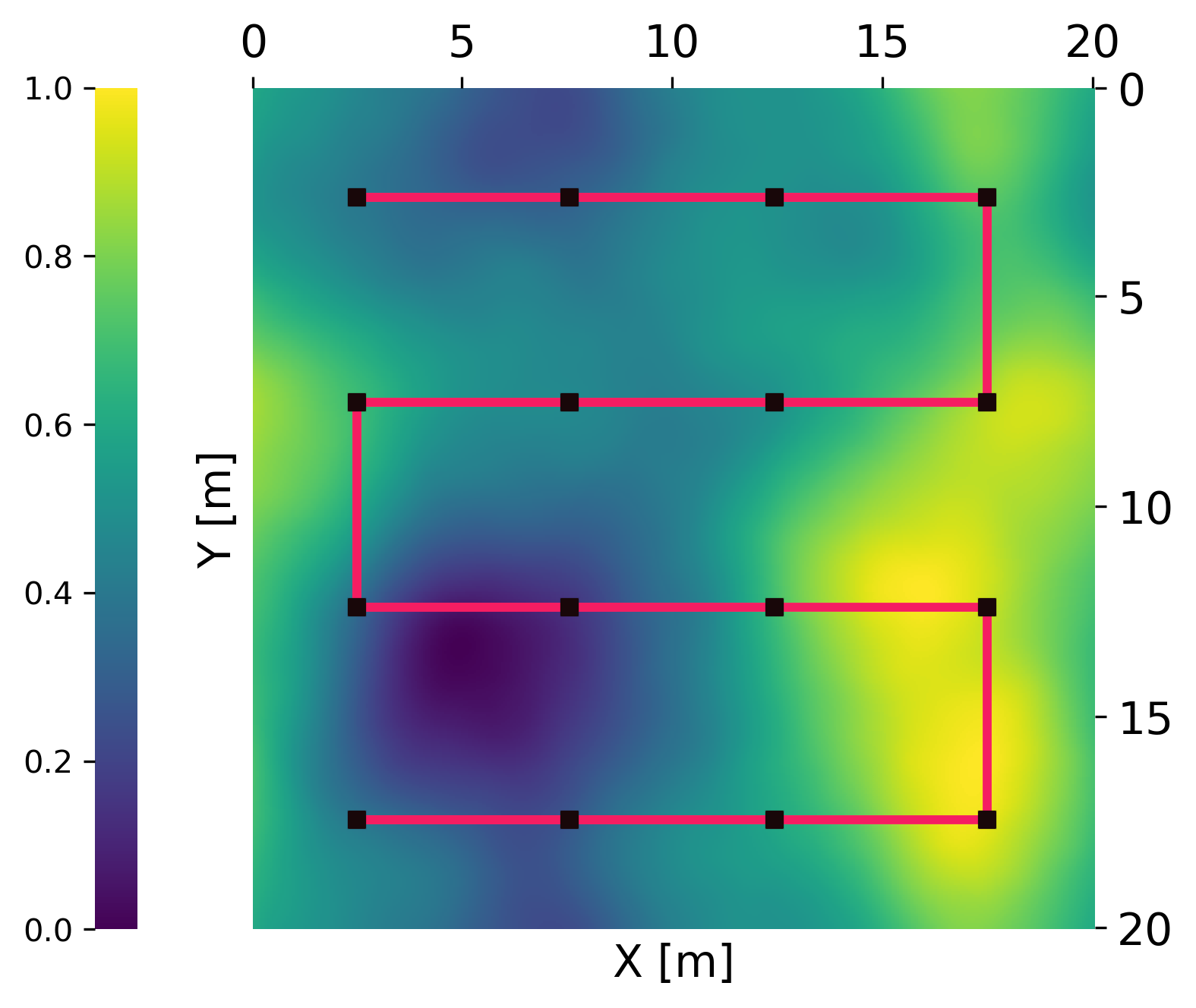}%
  \caption{Ground truth} \label{SF:mapping_results_gt}
\end{subfigure}
\begin{subfigure}{.38\columnwidth}
\includegraphics[width=\columnwidth]{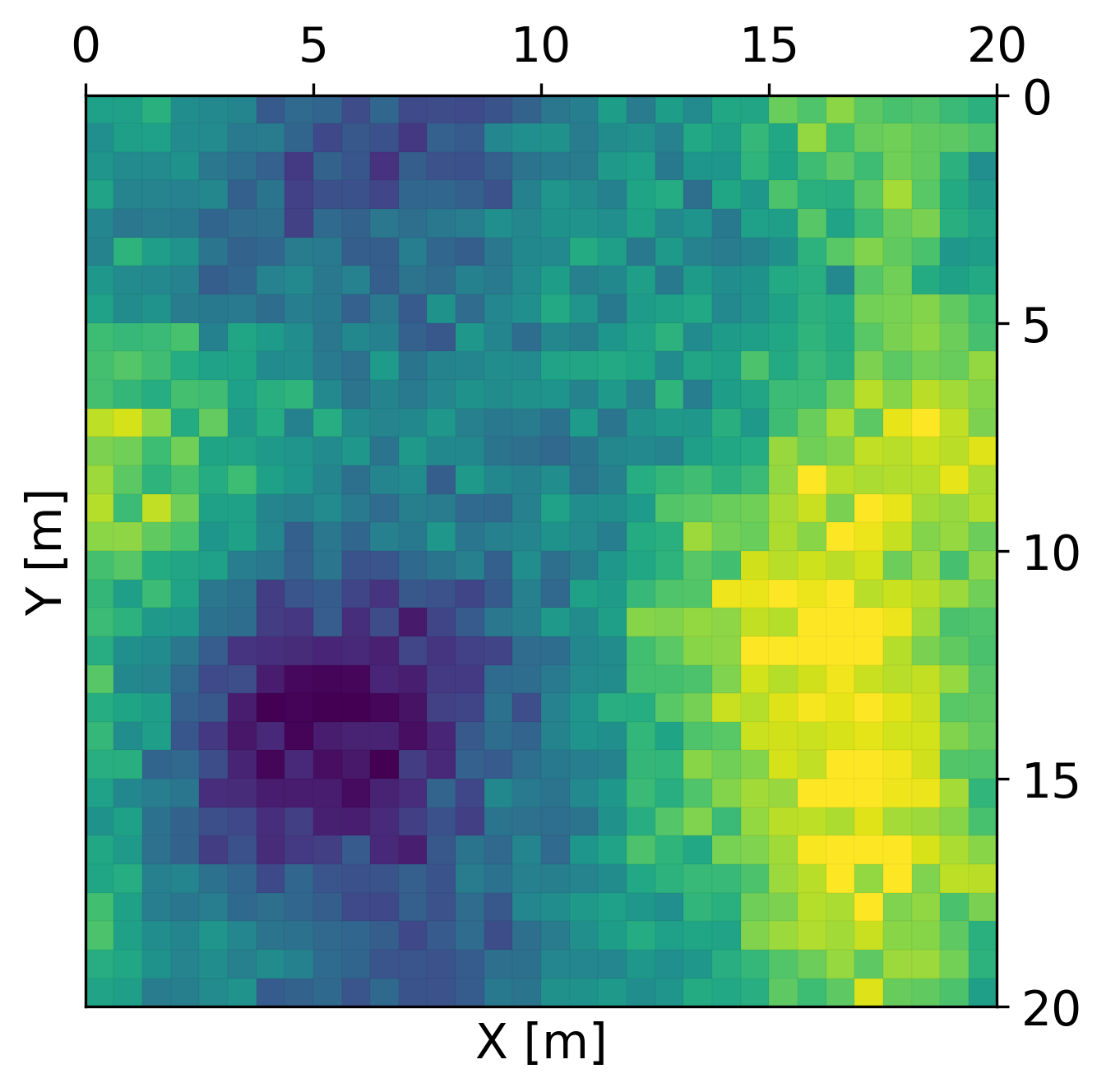}%
  \caption{FR-IDP}
\end{subfigure}
\begin{subfigure}{.38\columnwidth}
\includegraphics[width=\columnwidth]{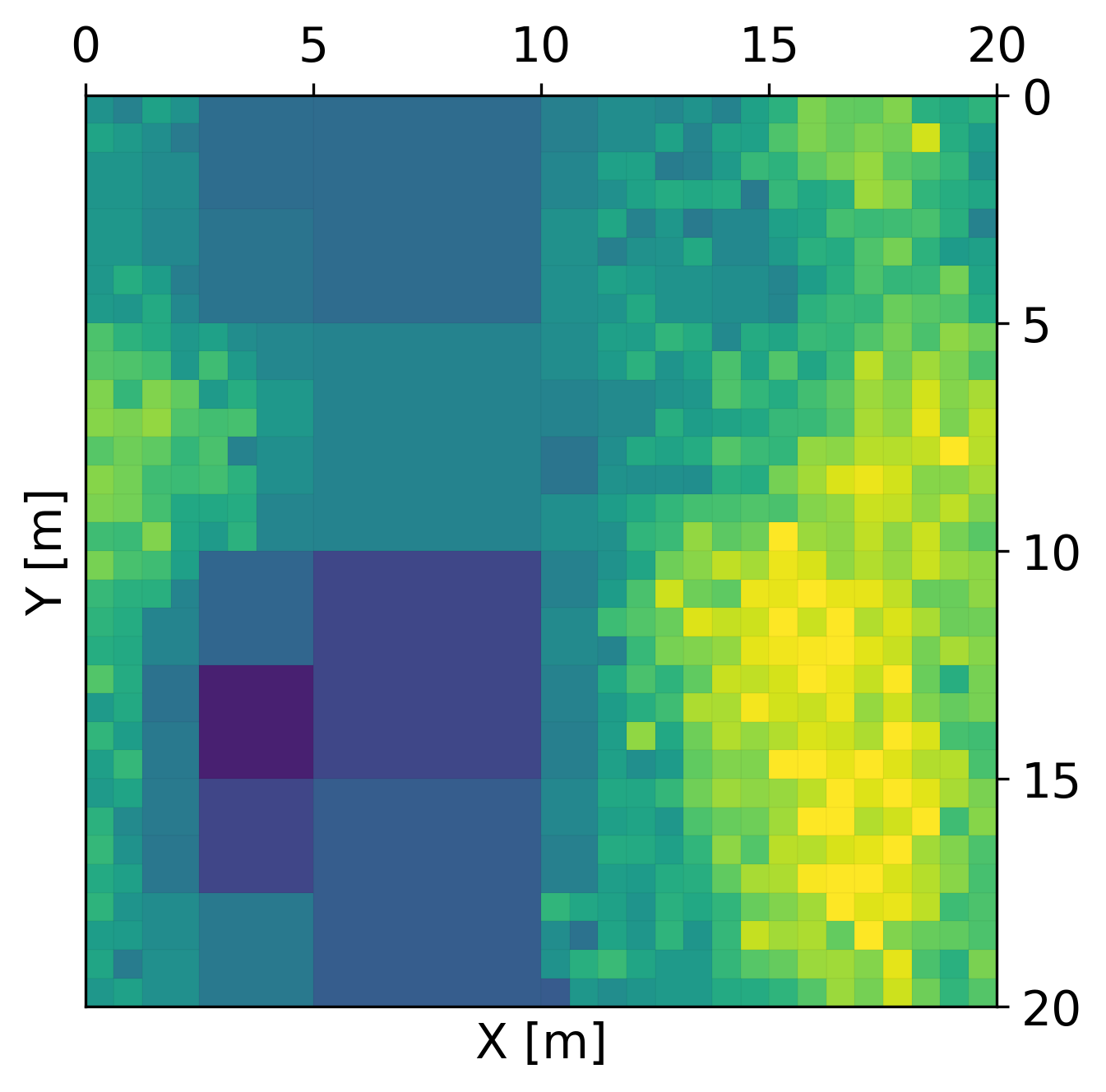}%
  \caption{AR-IDP}
\end{subfigure}
\begin{subfigure}{.38\columnwidth}
\includegraphics[width=\columnwidth]{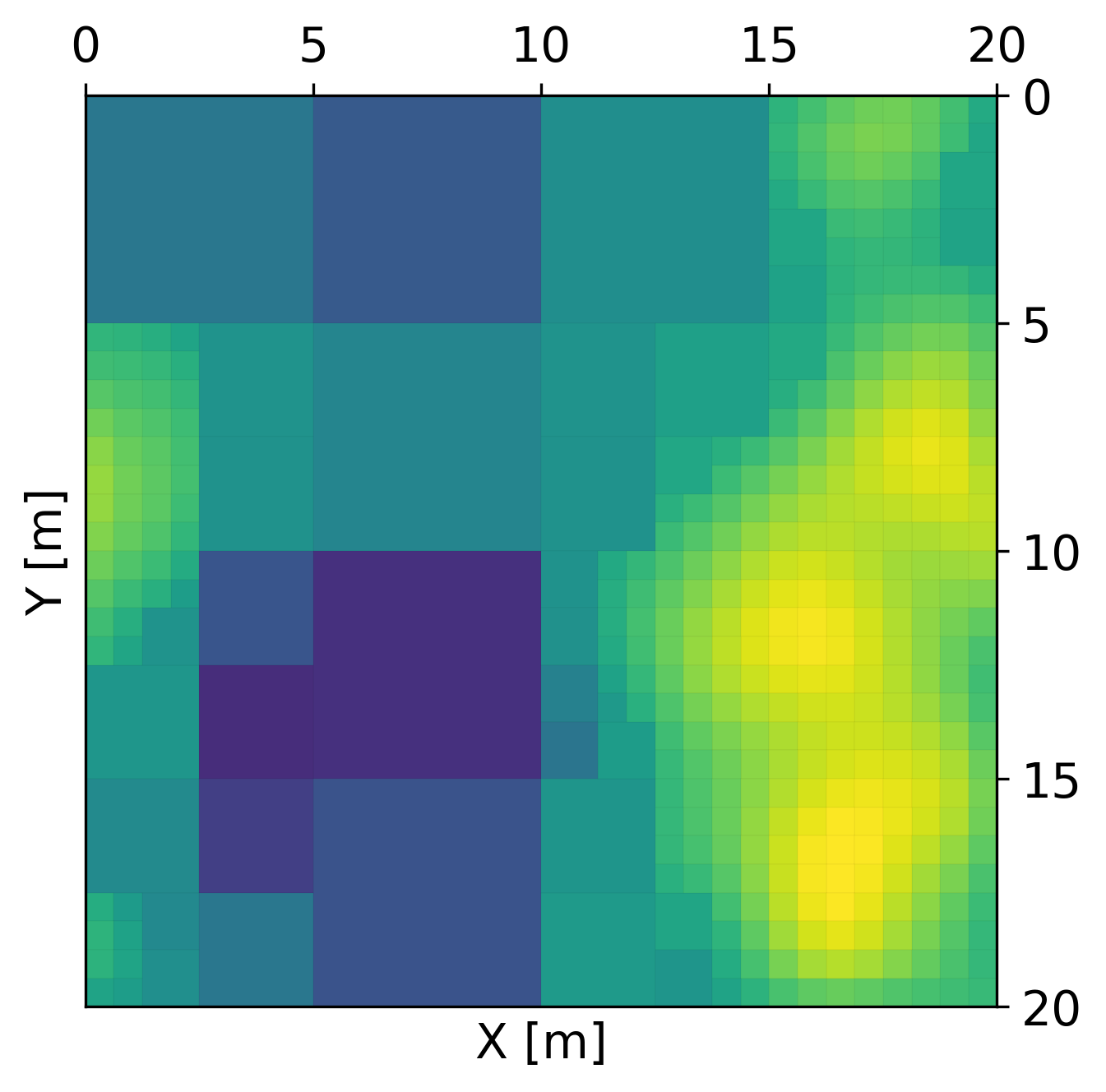}%
  \caption{AR-BCM}
\end{subfigure}

\begin{subfigure}{.38\columnwidth}
\includegraphics[width=\columnwidth]{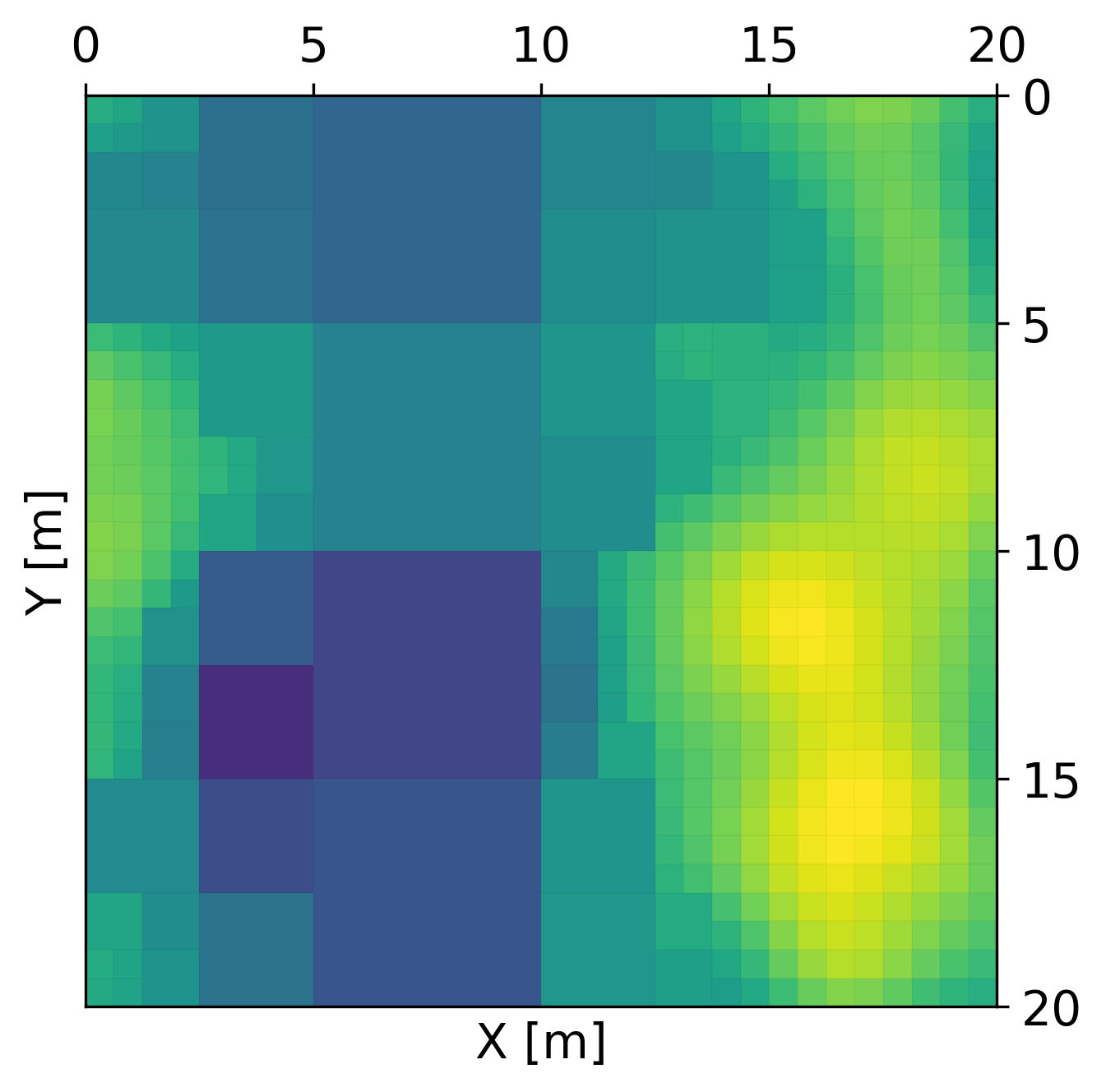}%
  \caption{AR-GPR-IK}
\end{subfigure}
\begin{subfigure}{.38\columnwidth}
\includegraphics[width=\columnwidth]{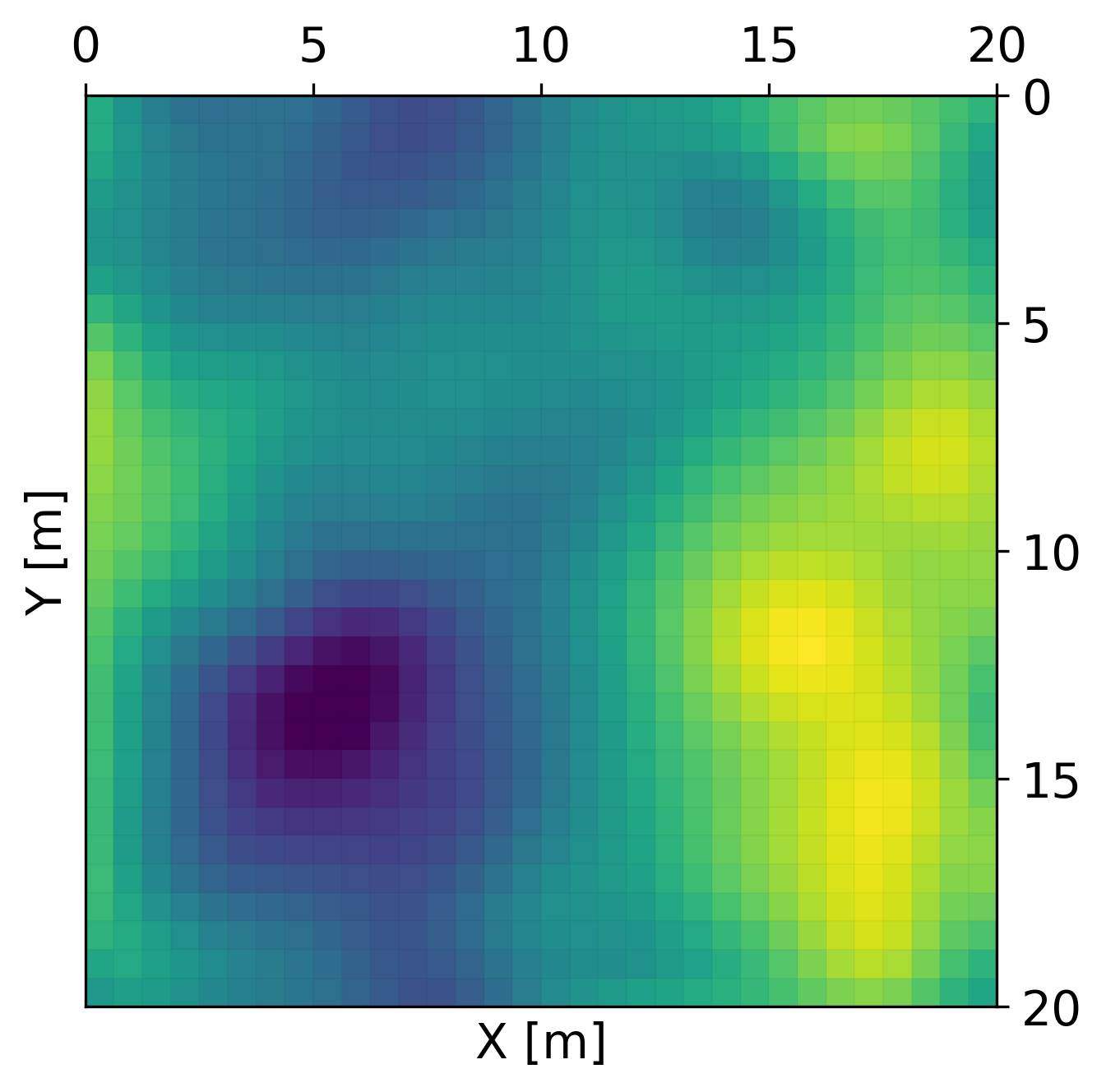}%
  \caption{FR-GPF}
\end{subfigure}
\begin{subfigure}{.38\columnwidth}
\includegraphics[width=\columnwidth]{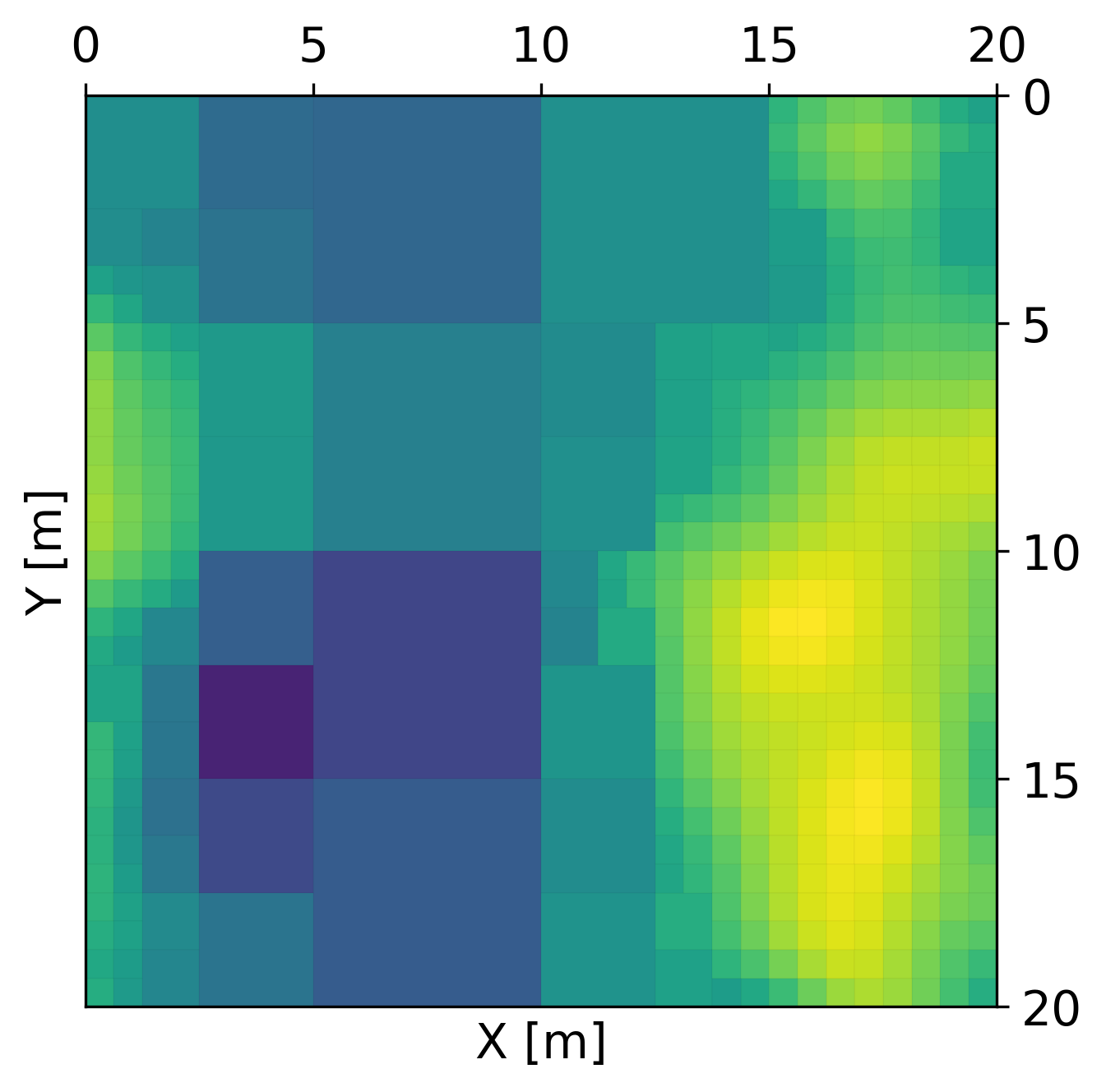}%
  \caption{Ours}
\end{subfigure}
\caption{Qualitative comparison of our approach (g) against benchmarks (b)-(f). The terrain is mapped using a lawnmower strategy, as shown in (a). The red line and black dots indicate the travelled path and measurement locations. All approaches use a map size of $32 \times 32$ grid cells.
By mapping adaptively, our method compresses information in areas with low information value (blue) while preserving details in higher-value areas of interest (yellow) to achieve an efficient, compact map representation for online applications.
}\label{F:mapping_results}
\end{figure*}

\section{Experimental Results} \label{S:experimental_results}
This section presents our experimental results. First, we evaluate our proposed mapping strategy by comparing it against different benchmarks in terrain mapping scenarios. Then, we validate our approach using real-world surface temperature data and integrate it into an adaptive path planning framework to demonstrate its benefits for online robotic applications.

\subsection{Mapping Evaluation} \label{SS:experiments_mapping}
We evaluate the mapping performance with total mapping time, mapping quality in terms of root mean square error (RMSE), intersection over union (IoU) of hotspots, memory consumption ratio and number of grid cells in the final maps. The total mapping time is calculated by aggregating the individual update times over the mapping task. RMSE and IoU are obtained by comparing the resulting maps with synthetic ground truth. We compare six different mapping approaches: 
\begin{itemize}
 \item \textit{FR-IDP}: vanilla fixed-resolution mapping under independence assumption~\citep{Elfes1989};
 \item \textit{AR-IDP}: adaptive-resolution mapping under independence assumption. Uninteresting grid cells are pruned during mapping as proposed in \citep{Einhorn2011};
 \item \textit{AR-BCM}: adaptive-resolution mapping using BCM and test-data tree, as adapted from \citep{Wang2016}. Uninteresting grid cells are pruned to reduce the number of query points in BCM. Note that we do not follow nested BCM approach as our whole map can be seen as a block in their case;
 \item \textit{AR-GPR-IK}: adaptive-resolution GP regression with integral kernel based on the original approach proposed in \citep{Reid2013}. We take one step further to recursively merge grid cells if they are uninteresting after each regression update;
 \item \textit{FR-GPF}: fixed-resolution GP fusion proposed in \citep{Popovic2017a};
 \item \textit{Ours}: our new adaptive-resolution mapping strategy based on GP fusion with integral kernel, as described in \cref{S:approach}.
\end{itemize}

\begin{table*}[h]
\centering
\begin{tabular}{llllllll}
             \makecell[l]{Map \\ size} & \makecell[l]{Method} & \makecell[l]{RMSE $\downarrow$} & \makecell[l]{RMSE \\ (hotspots) $\downarrow$} & \makecell[l]{IoU \\ (hotspots) $\uparrow$} & \makecell[l]{Mapping \\ time [ms] $\downarrow$} & \makecell[l]{Memory \\ usage ratio [\%] $\downarrow$} & \makecell[l]{Number of \\ map cells $\downarrow$} \\\midrule[1.2pt]
$16\times16$  
 & FR-IDP     &$0.045 \pm 0.002$   &$0.045\pm 0.002$     & $0.813\pm 0.024$     &$5.575\pm 0.466$    &$4.187\pm 0$    & $256\pm 0$      \\
 & AR-IDP     &$0.071 \pm 0.003$   &$0.046\pm 0.002$     & $0.812\pm 0.024$     &$7.299\pm 0.901$    &$2.155\pm 1.127$    & $125.2\pm 18.258$      \\
 & AR-BMC     &$0.071 \pm 0.003$   &$0.038\pm 0.003$       & $0.856\pm 0.023$  &$273.546\pm 69.549$   &$49.892\pm 11.433$    & $115.22\pm 16.167$      \\
 & AR-GPR-IK  &$0.065 \pm 0.003$   &$0.037\pm 0.002$       & $0.856\pm 0.024$  &$59.379\pm 20.680$    &$40.692\pm 8.282$    & $118.62\pm 16.907$      \\
 & FR-GPF     &$0.037 \pm 0.002$     &$0.037\pm 0.002$       & $0.857\pm 0.023$  &$11.375\pm 0.965$     &$100\pm 0$    & $256\pm0$      \\
 & Ours  &$0.065 \pm 0.003$     &$0.037\pm 0.002$       & $0.857\pm 0.026$  &$10.168\pm 1.119$     &$38.729\pm 9.416$    & $114.4\pm 19.754$      \\
  
  \midrule
$32\times32$ 
 & FR-IDP     &$0.065 \pm 0.004$     &$0.065 \pm 0.004$       & $0.723\pm 0.021$  &$28.968\pm 0.834$              &$1.067\pm 0$    & $1024\pm 0$      \\
 & AR-IDP     &$0.079 \pm 0.005$     &$0.067 \pm 0.003$       & $0.725\pm 0.020$  &$63.796\pm 6.880$              &$0.529\pm 0.068$    & $508\pm 65.121$      \\
 & AR-BMC     &$0.071 \pm 0.005$     &$0.025 \pm 0.003$       & $0.864\pm 0.023$  &$13100.225\pm 2009.709$              &$26.433\pm 5.359$    & $356.25\pm 74.012$      \\
 & AR-GPR-IK  &$0.066 \pm 0.004$     &$0.026 \pm 0.003$       & $0.866\pm 0.024$  &$1747.631\pm 677.670$              &$17.538\pm 5.640$    & $371.3\pm 74.279$      \\
 & FR-GPF     &$0.027 \pm 0.002$     &$0.026 \pm 0.002$       & $0.867\pm 0.024$  &$430.843\pm 7.447$              &$100\pm 0$    & $1024\pm 0$      \\
 & Ours  &$0.065 \pm 0.004$     &$0.026 \pm 0.003$       & $0.867\pm 0.025$  &$261.763\pm 24.443$              &$16.632\pm 5.112$    & $360.4\pm 71.003$      \\
  
 \midrule
$64\times64$ 
 & FR-IDP     &$0.123 \pm 0.003$     &$0.123\pm 0.008$       & $0.625\pm 0.056$  &$123.562\pm 4.271$             &$0.268\pm 0$    & $4096\pm 0$      \\
 & AR-IDP     &$0.127 \pm 0.004$     &$0.124\pm 0.008$       & $0.623\pm 0.058$  &$586.427\pm 43.362$            &$0.218\pm 0.023$    & $2687.14\pm 383.364$      \\
 & AR-BMC     &$0.104 \pm 0.004$    &$0.025 \pm 0.003$       & $0.869 \pm 0.028$ &$68645.365 \pm4606.239$       &$15.782\pm 4.125$  & $1258.5 \pm473.78$      \\
 & AR-GPR-IK  &$0.073 \pm 0.004$     &$0.025\pm 0.003$       & $0.872\pm 0.021$  &$18620.558\pm 6448.663$     &$11.552\pm 4.221$    & $1387\pm 443.744$      \\
 & FR-GPF     &$0.024 \pm 0.002$     &$0.024\pm 0.002$       & $0.875\pm 0.023$  &$9977.749\pm 251.003$       &$100\pm 0$    & $4096\pm 0$      \\
 & Ours  &$0.073 \pm 0.004$     &$0.024\pm 0.002$       & $0.876\pm 0.023$  &$4098.029\pm 548.881$       &$8.877\pm 4.824$    & $1271.25\pm 484.191$      \\
\end{tabular}
\caption{Comparison of our approach against benchmarks for varying map sizes. By combining GP fusion and adaptive-resolution mapping using an integral kernel, our strategy reduces runtime while delivering highly accurate maps. Note that memory usage is reported as a ratio relative to the fixed-resolution GP fusion (FR-GPF) approach.} \label{T:results}
\end{table*}

We simulate 20 $20m \times 20m$ Gaussian random fields as ground truth environments representing a spatially-correlated field on a terrain. 
For simplification, the ground truth field values are normalized to $[0, 1]$ and we define regions with values $> 0.7$ as hotspots of interest. To assess mapping performance at different scales, we conduct experiments at 3 different maximal resolutions: $16 \times 16$, $32 \times 32$, and $64 \times 64$ grid cell maps corresponding to adaptive-resolution approaches with maximum tree depths of 4, 5, and 6, respectively, in a quadtree configuration. Note that a quadtree is a special case of an ND-tree, thus different mapping settings are possible by adapting the general ND-tree decomposition.  

The terrains are mapped using a lawnmower pattern to focus on comparing the methods in terms of mapping performance only, excluding the influence of path variations. To simulate a UAV monitoring mission, we take 16 non-overlapping measurements as shown in \cref{SF:mapping_results_gt} to fully cover the terrain, assuming a flight altitude of $2.5m$ and $5m \times 5m$ sensor footprint on the ground. For all GP-based mapping approaches (\textit{AR-BCM}, \textit{AR-GPR-IK}, \textit{FR-GPF} and \textit{Ours}), the SE kernel function with hyperparameters $\bm{\theta} = \{\sigma^2 ,\,l\} = \{1,\,2.36\}$ and constant prior mean of 0.5 are applied. For approaches using an integral kernel (\textit{AR-GPR-IK}, \textit{Ours}), we follow \cref{E:average_mean,E:average_kernel} to calculate the prior maps. For mapping under independence assumption (\textit{FR-IDP}, \textit{AR-IDP}), we use the same prior mean and variance, while setting all cross-correlation terms to $0$ to isolate each grid cell. We consider the `average measurements' model in all mapping approaches. For merging cells in adaptive-resolution approaches, we choose $\{\gamma,\, f_\text{th}\} = \{2,0.7\}$ in \cref{E: hotspots}. Note that all these hyperparameters are manually tuned but used consistently in all experiments.

The results are summarized in \cref{T:results} and \cref{F:mapping_results}. In all cases, approaches relying on the cell independence assumption yield least accurate maps with highest RMSE and lowest IoU, since they are most vulnerable to sensor noise or sparse measurements. This is because they neglect correlations for mapping, which are key for capturing continuous variables. In contrast, the four GP-based approaches reflect the smooth structure of the Gaussian random field, as they incorporate covariance information into the map update. As expected, the averaging effect caused by merging cells in adaptive-resolution approaches leads to higher total RMSE compared to \textit{FR-GPF}. However, all GP-based approaches show the same accuracy in mapping hotspots as well as IoU scores, as required in our problem setup. In terms of mapping efficiency, \textit{AR-BCM} performs the worst as it executes large matrix inversion and BCM fusion at every update step, leading to prohibitively slow mapping. Note that BCM benefits from parallelizing several GP regressions. However, in online mapping scenarios, where measurements are accumulated incrementally, BCM loses this strength. \textit{AR-GPR-IK} is slower than two GP fusion approaches (\textit{FR-GPF} and \textit{Ours}), due to regression using accumulated measurements. We point out that by using the integral kernel together with the average measurement model, \textit{AR-GPR-IK} already achieves significant speed-up compared to vanilla GP regression. In all cases, \textit{AR-IDP} is slower than \textit{FR-IDP} due to overhead caused by tree search. The same overhead is expected in our approach, however, as the major bottleneck in fusion is the matrix inversion and multiplication in \cref{E:kf_mean,E:kf_cov}, this can be compensated by faster Bayesian fusion update with less grid cells in our approach. In terms of memory usage, \textit{FR-GPF} consumes the most memory space as it maintains a large constant number of grid cells and a large covariance matrix. Among the adaptive-resolution approaches, \textit{AR-IDP} shows the worst merging ability, as indicated by the number of grid cells in the final map. This can be explained by heterogeneous states in children nodes caused by inaccurate mapping, which potentially reduces the chance of merging operation. Among the GP-based approaches, our new approach achieves the fastest mapping updates and best memory compression ratios with competitive map quality. In all cases, our approach outperforms \textit{AR-IDP} and \textit{FR-IDP} in terms of map quality.
\subsection{Validation on Real-World Data}
\begin{figure}[h]
\centering
  \begin{subfigure}[]{0.4\textwidth}
  \includegraphics[width=\textwidth, height=3cm]{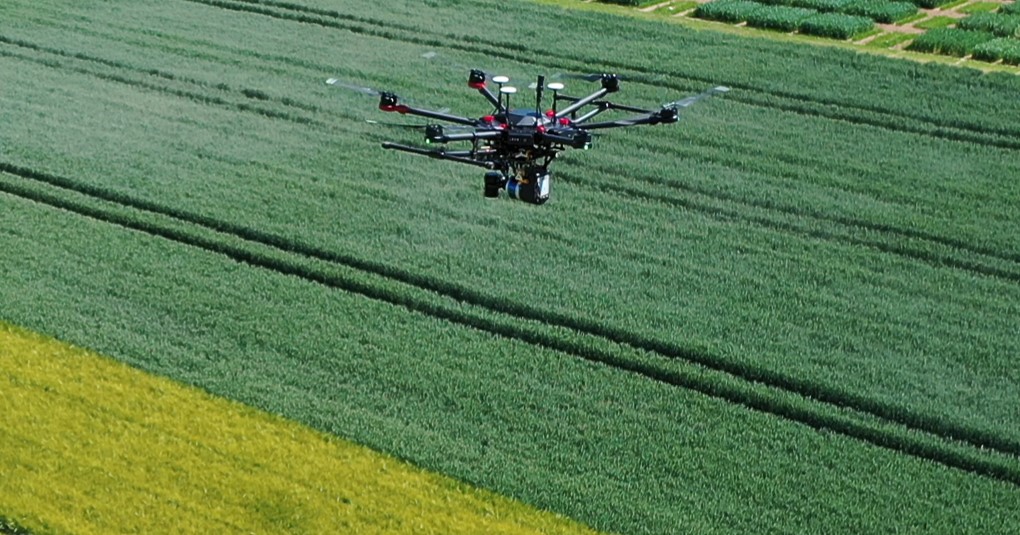}
  \end{subfigure}
  \begin{subfigure}[]{0.23\textwidth}
  \includegraphics[width=\textwidth]{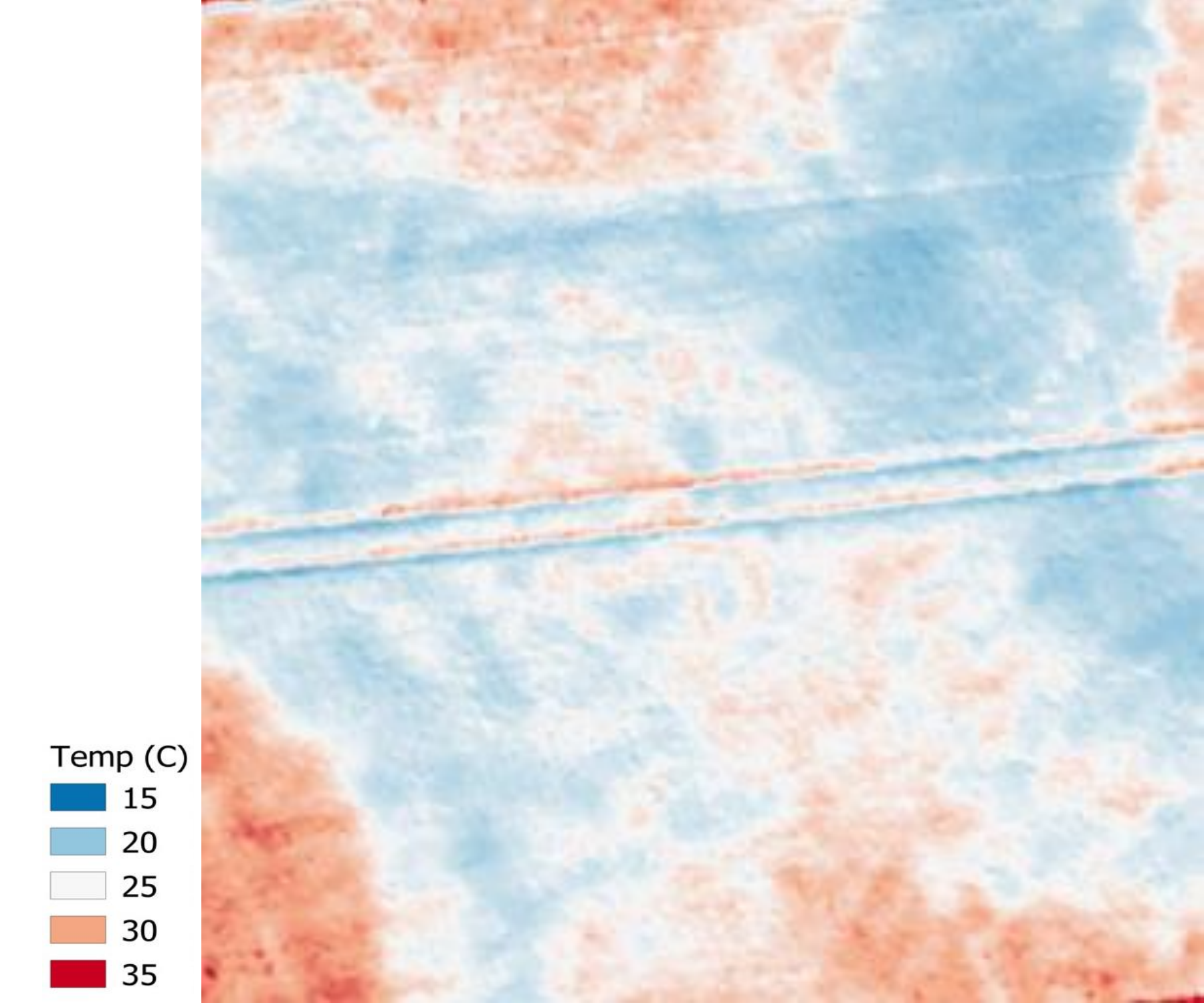}
  \end{subfigure}\hfill%
  \begin{subfigure}[]{0.235\textwidth}
  \includegraphics[width=\textwidth]{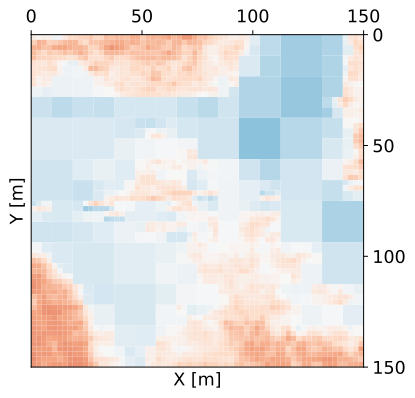}
  \end{subfigure}\hfill%
    \caption{Validation of our approach for surface temperature mapping. Top: Experimental setup showing our UAV over the crop field. Bottom-left: temperature data of a crop field. Bottom-right: result of mapping this area using our method. High-temperature areas of interest (red) are mapped at higher resolutions to preserve detail in these regions. }\label{F:thermal_field}
\end{figure}

We demonstrate our mapping approach using real-world surface temperature data. The data was collected in a $150m \times 150m$ crop field $(50.86\degree\,\textrm{lat.}, 6.45\degree\, \textrm{lon.})$ near J\"{u}lich, Germany on June 25, 2021 using a DJI Matrice 600 UAV platform equipped with a Vue Pro R 640 thermal sensor. During data acquisition, the UAV followed a lawnmower path at $100m$ altitude to collect images at $15cm$ spatial ground resolution. The images were processed using Pix4D software to create an orthomosaic used as a proxy for ground truth in our experiment. We use a maximal map resolution of $64 \times 64$. The entire mapping takes $28.31s$ considering a $81$ measurements with $50\%$ overlap. The aim is to validate our method for adaptively mapping temperature hotspots ($> 28^{\circ}$C) at finer resolutions using this real data. The mapping result in \cref{F:thermal_field} confirms that our approach can adapt the map resolution in a targeted way. For mapping at larger scales or at higher resolutions, our approach could be used to generate local sub-maps which are then fused by a BCM or the approach proposed by \citet{Sun2015}.

\subsection{Application for Adaptive Path Planning}
Finally, we integrate our mapping approach into an adaptive path planning framework for UAV-based terrain monitoring~\citep{Popovic2020} to demonstrate its applicability for online robotic scenarios. The goal of the planning task is to efficiently detect regions of interest in an initially unknown environment under time constraints. For this, the UAV must adaptively plan its path based on the current map to trade off between exploration and exploitation. In this experiment, we consider the same setup as described in \Cref{SS:experiments_mapping} except setting our prior mean to $0.7$ to initially encourage exploration. We compare the \textit{FR-IDP}, \textit{AR-IDP}, \textit{FR-GPF} methods to our approach, 
as regression-based mapping approaches prohibitively slow for online planning. We use a 3D lattice consisting of $300$ total waypoints at altitudes of $2m$ and $5m$ to represent the discrete action space. The planner applies greedy search among these candidates to find the next best measurement position by forward-simulating the map update and calculating the reward. The information-theoretic reward is defined by the posterior variance reduction in regions of interest divided by the flight time to the waypoint candidate. 
For more technical details, please refer to the planning framework of~\citet{Popovic2020}.

\begin{figure}[h]
\centering
\begin{subfigure}{\columnwidth}
\includegraphics[width=8cm, height=4cm]{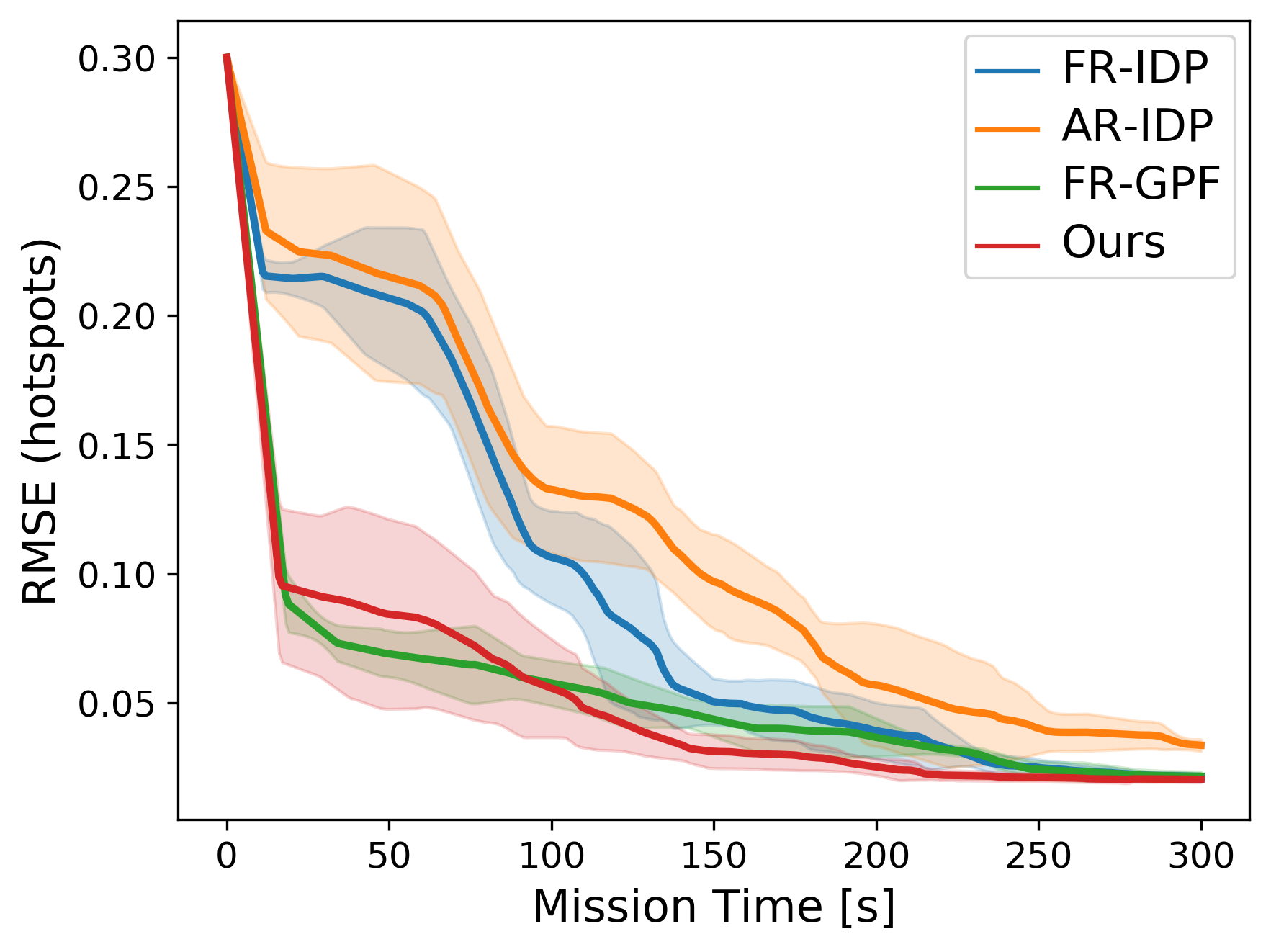}%
\end{subfigure}
\begin{subfigure}{\columnwidth}
\includegraphics[width=8cm, height=4cm]{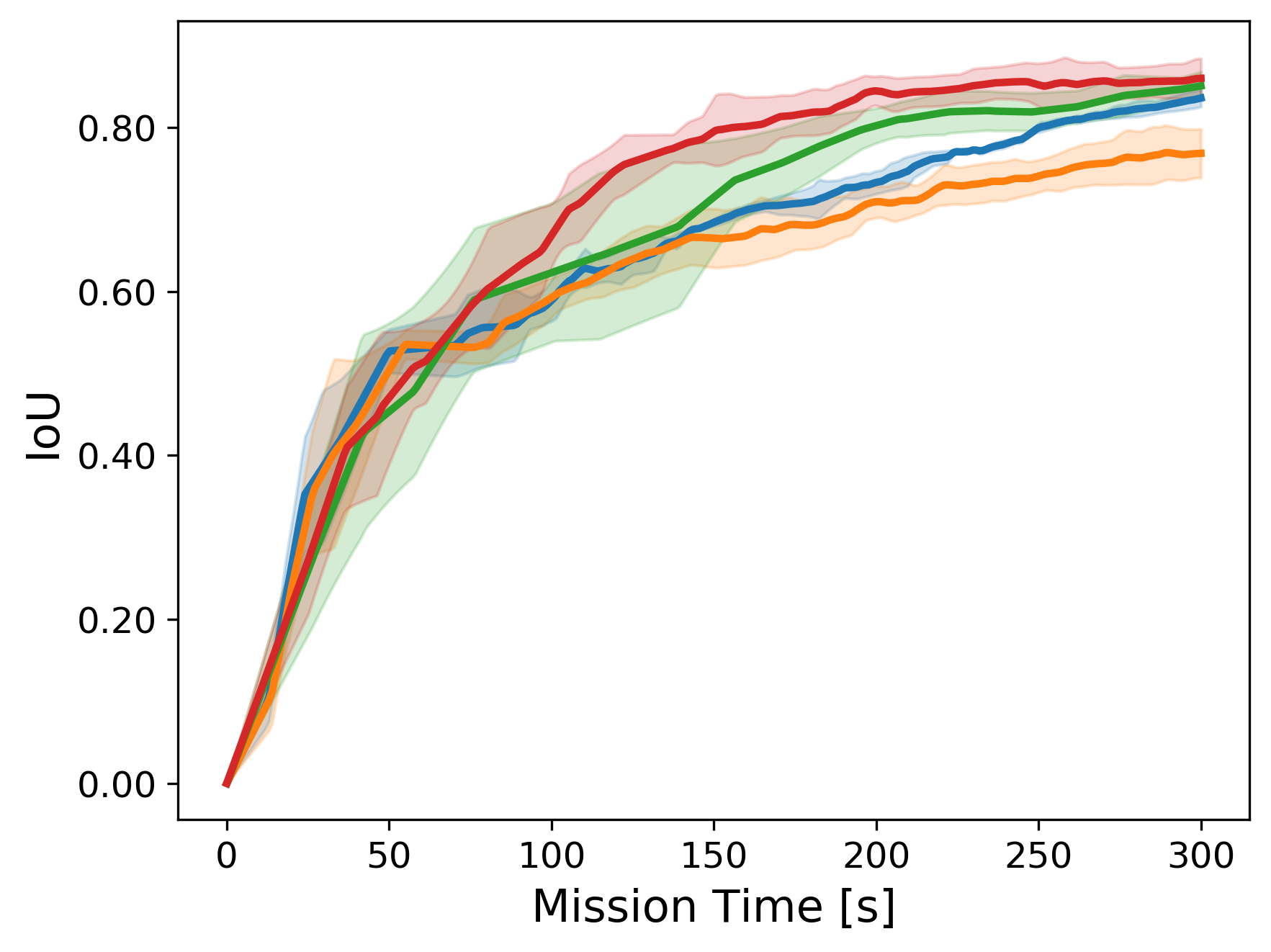}%
\end{subfigure}
 \caption{Comparison of mapping approaches for adaptive path planning in terrain monitoring scenarios. Our strategy (red) performs best to efficiently reconstruct hotspot regions in an unknown environment with highest mapping accuracy (top) and map quality (bottom). Solid lines represent means over 10 trials and shaded regions indicate standard deviations.}\label{F:planning}
\end{figure}

We conduct experiments on 10 simulated Gaussian random fields and plot the evolution of RMSE in hotspots and IoU over mission time in \cref{F:planning}. The mission time is composed of planning time, map update time, and flight time.
The results show that planning using our mapping approach achieves the best IoU and RMSE (hotspots) scores with the shortest mission time, which is favorable for autonomous monitoring tasks using resource-constrained UAVs.
Planning using our approach outperforms \textit{FR-GPF} due to more efficient map updates, which significantly accelerates forward-simulation during predictive planning.
The poorer planning performance using \textit{FR-IDP} and \textit{AR-IDP} is a direct consequence of inaccurate mapping results. 
As observed in \cref{SS:experiments_mapping}, mapping approaches using independence assumption neglect important spatial correlation and are thus more susceptible to sensor noise.
At the start of the mission, the UAV explores the unknown environment at a higher altitude, as quickly covering the unexplored areas results in the highest reward. Note that measurements from higher altitudes are noisier, therefore severely degrading the map quality. After exploration, the UAV focuses on the observed hotspots. However, due to inaccurate mapping from previous measurements, the false positive interesting areas mislead the UAV, leading to the close inspection of actually uninteresting regions. This inaccuracy deprives \textit{FR-IDP} and \textit{AR-IDP} of their advantage in fast planning. 

\section{Conclusions and Future Work} \label{S:conclusions_and_future_work}
In this paper, we propose a new approach for online field mapping. We introduce a novel use of an integral kernel in GP fusion framework and use ND-tree to store our map. Combining these two elements enables us to adapt the map resolution on-the-fly while neatly maintaining spatial correlations. Results show that our approach achieves competitive performance in terms of mapping efficiency, memory usage, and map quality. The applicability of our approach is validated using real-world data in a surface temperature mapping scenario. Moreover, we demonstrate that faster and more accurate map updates facilitate adaptive path planning for efficient information gathering in robotic applications.

A major drawback of our approach is the irreversible merging operation, which could limit its applicability in dynamic environments, as merged grid cells cannot be efficiently partitioned again. Future work will study adaptive-resolution mapping approach for large-scale and dynamic environments.

\section*{Acknowledgement}
We would like to thank Mr. Jordan Bates from Forschungszentrum J\"{u}lich for providing the temperature data.

\bibliographystyle{IEEEtranN}
\footnotesize
\bibliography{2022-ral-jin}

\begin{thebibliography}{29}
\providecommand{\natexlab}[1]{#1}
\providecommand{\url}[1]{#1}
\csname url@samestyle\endcsname
\providecommand{\newblock}{\relax}
\providecommand{\bibinfo}[2]{#2}
\providecommand{\BIBentrySTDinterwordspacing}{\spaceskip=0pt\relax}
\providecommand{\BIBentryALTinterwordstretchfactor}{4}
\providecommand{\BIBentryALTinterwordspacing}{\spaceskip=\fontdimen2\font plus
\BIBentryALTinterwordstretchfactor\fontdimen3\font minus
  \fontdimen4\font\relax}
\providecommand{\BIBforeignlanguage}[2]{{%
\expandafter\ifx\csname l@#1\endcsname\relax
\typeout{** WARNING: IEEEtranN.bst: No hyphenation pattern has been}%
\typeout{** loaded for the language `#1'. Using the pattern for}%
\typeout{** the default language instead.}%
\else
\language=\csname l@#1\endcsname
\fi
#2}}
\providecommand{\BIBdecl}{\relax}
\BIBdecl

\bibitem[Dunbabin and Marques(2012)]{Dunbabin2012}
M.~Dunbabin and L.~Marques, ``{Robots for Environmental Monitoring: Significant
  Advancements and Applications},'' \emph{IEEE Robotics \& Automation
  Magazine}, vol.~19, no.~1, pp. 24--39, 2012.

\bibitem[Manfreda et~al.(2018)Manfreda, McCabe, Miller, Lucas,
  Pajuelo~Madrigal, Mallinis, Ben~Dor, Helman, Estes, Ciraolo,
  et~al.]{Manfreda2018}
S.~Manfreda, M.~F. McCabe, P.~E. Miller, R.~Lucas, V.~Pajuelo~Madrigal,
  G.~Mallinis, E.~Ben~Dor, D.~Helman, L.~Estes, G.~Ciraolo \emph{et~al.}, ``{On
  the Use of Unmanned Aerial Systems for Environmental Monitoring},''
  \emph{Remote Sensing}, vol.~10, no.~4, 2018.

\bibitem[Tmušić et~al.(2020)Tmušić, Manfreda, Aasen, James, Gonçalves,
  Ben-Dor, Brook, Polinova, Arranz, Mészáros, Zhuang, Johansen, Malbeteau,
  de~Lima, Davids, Herban, and McCabe]{Tmusic2020}
G.~Tmušić, S.~Manfreda, H.~Aasen, M.~R. James, G.~Gonçalves, E.~Ben-Dor,
  A.~Brook, M.~Polinova, J.~J. Arranz, J.~Mészáros, R.~Zhuang, K.~Johansen,
  Y.~Malbeteau, I.~P. de~Lima, C.~Davids, S.~Herban, and M.~F. McCabe,
  ``{Current Practices in UAS-based Environmental Monitoring},'' \emph{Remote
  Sensing}, vol.~12, no.~6, 2020.

\bibitem[Popovi\'{c} et~al.(2017)Popovi\'{c}, Vidal-Calleja, Hitz, Sa,
  Siegwart, and Nieto]{Popovic2017a}
M.~Popovi\'{c}, T.~Vidal-Calleja, G.~Hitz, I.~Sa, R.~Siegwart, and J.~Nieto,
  ``{Multiresolution mapping and informative path planning for UAV-based
  terrain monitoring},'' in \emph{IEEE/RSJ International Conference on
  Intelligent Robots and Systems}, 2017, pp. 1382--1388.

\bibitem[Hollinger and Sukhatme(2014)]{Hollinger2014}
G.~A. Hollinger and G.~S. Sukhatme, ``Sampling-based robotic information
  gathering algorithms,'' \emph{The International Journal of Robotics
  Research}, vol.~33, pp. 1271 -- 1287, 2014.

\bibitem[Stache et~al.(2021)Stache, Westheider, Magistri, Popović, and
  Stachniss]{Stache2021}
F.~Stache, J.~Westheider, F.~Magistri, M.~Popović, and C.~Stachniss,
  ``Adaptive path planning for uav-based multi-resolution semantic
  segmentation,'' in \emph{European Conference on Mobile Robots}, 2021.

\bibitem[Bircher et~al.(2016)Bircher, Kamel, Alexis, Burri, Oettershagen,
  Omari, Mantel, and Siegwart]{Bircher2016}
A.~Bircher, M.~Kamel, K.~Alexis, M.~Burri, P.~Oettershagen, S.~Omari,
  T.~Mantel, and R.~Siegwart, ``Three-dimensional coverage path planning via
  viewpoint resampling and tour optimization for aerial robots,''
  \emph{Autonomous Robots}, vol.~40, pp. 1059--1078, 2016.

\bibitem[Hornung et~al.(2013)Hornung, Wurm, Bennewitz, Stachniss, and
  Burgard]{Hornung2013}
A.~Hornung, K.~M. Wurm, M.~Bennewitz, C.~Stachniss, and W.~Burgard, ``{OctoMap:
  An efficient probabilistic 3D mapping framework based on octrees},''
  \emph{Autonomous Robots}, vol.~34, no.~3, pp. 189--206, 2013.

\bibitem[Funk et~al.(2021)Funk, Tarrio, Papatheodorou, Popovi\'{c},
  Alcantarilla, and Leutenegger]{Funk2021}
N.~Funk, J.~Tarrio, S.~Papatheodorou, M.~Popovi\'{c}, P.~F. Alcantarilla, and
  S.~Leutenegger, ``{Multi-Resolution {3D} Mapping with Explicit Free Space
  Representation for Fast and Accurate Mobile Robot Motion Planning},''
  \emph{Robotics and Automation Letters}, vol.~6, pp. 3553--3560, 2021.

\bibitem[Einhorn et~al.(2011)Einhorn, Schr{\"{o}}ter, and Gross]{Einhorn2011}
E.~Einhorn, C.~Schr{\"{o}}ter, and H.~M. Gross, ``{Finding the adequate
  resolution for grid mapping - Cell sizes locally adapting on-the-fly},'' in
  \emph{IEEE International Conference on Robotics and Automation}.\hskip 1em
  plus 0.5em minus 0.4em\relax IEEE, 2011, pp. 1843--1848.

\bibitem[Vidal-Calleja et~al.(2014)Vidal-Calleja, Su, De~Bruijn, and
  Miro]{Teresa2014}
T.~Vidal-Calleja, D.~Su, F.~De~Bruijn, and J.~V. Miro, ``{Learning spatial
  correlations for Bayesian fusion in pipe thickness mapping},'' in \emph{IEEE
  International Conference on Robotics and Automation}.\hskip 1em plus 0.5em
  minus 0.4em\relax IEEE, 2014, pp. 683--690.

\bibitem[Popovi\'{c} et~al.(2020)Popovi\'{c}, Vidal-Calleja, Hitz, Chung, Sa,
  Siegwart, and Nieto]{Popovic2020}
M.~Popovi\'{c}, T.~Vidal-Calleja, G.~Hitz, J.~J. Chung, I.~Sa, R.~Siegwart, and
  J.~Nieto, ``{An informative path planning framework for UAV-based terrain
  monitoring},'' \emph{Autonomous Robots}, vol.~44, pp. 889--911, 2020.

\bibitem[Reece and Roberts(2010)]{Reece2010}
S.~Reece and S.~Roberts, ``{An introduction to Gaussian processes for the
  Kalman filter expert},'' in \emph{International Conference on Information
  Fusion}.\hskip 1em plus 0.5em minus 0.4em\relax IEEE, 2010.

\bibitem[Stachniss et~al.(2009)Stachniss, Plagemann, and
  Lilienthal]{Stachniss2009}
C.~Stachniss, C.~Plagemann, and A.~Lilienthal, ``{Learning Gas Distribution
  Models using Sparse Gaussian Process Mixtures},'' \emph{Autonomous Robots},
  vol.~26, pp. 187--202, 2009.

\bibitem[Hitz et~al.(2014)Hitz, Gotovos, Pomerleau, Garneau, Pradalier, Krause,
  and Siegwart]{Gregory2014}
G.~Hitz, A.~Gotovos, F.~Pomerleau, M.-E. Garneau, C.~Pradalier, A.~Krause, and
  R.~Siegwart, ``{Fully Autonomous Focused Exploration for Robotic
  Environmental Monitoring},'' in \emph{IEEE International Conference on
  Robotics and Automation}.\hskip 1em plus 0.5em minus 0.4em\relax IEEE, 2014,
  pp. 2658--2664.

\bibitem[Hitz et~al.(2017)Hitz, Galceran, Garneau, Pomerleau, and
  Siegwart]{Hitz2017}
G.~Hitz, E.~Galceran, M.~{\`{E}}. Garneau, F.~Pomerleau, and R.~Siegwart,
  ``{Adaptive continuous-space informative path planning for online
  environmental monitoring},'' \emph{Journal of Field Robotics}, vol.~34,
  no.~8, pp. 1427--1449, 2017.

\bibitem[Vasudevan et~al.(2009)Vasudevan, Ramos, Nettleton, Durrant-Whyte, and
  Blair]{Shrihari2009}
S.~Vasudevan, F.~Ramos, E.~Nettleton, H.~Durrant-Whyte, and A.~Blair,
  ``{Gaussian Process modeling of large scale terrain},'' in \emph{IEEE
  International Conference on Robotics and Automation}.\hskip 1em plus 0.5em
  minus 0.4em\relax IEEE, 2009, pp. 1047--1053.

\bibitem[Elfes(1989)]{Elfes1989}
A.~Elfes, ``Using occupancy grids for mobile robot perception and navigation,''
  \emph{Computer}, vol.~22, no.~6, pp. 46--57, 1989.

\bibitem[O’Callaghan and Ramos(2012)]{Simon2012}
S.~T. O’Callaghan and F.~T. Ramos, ``Gaussian process occupancy maps,''
  \emph{The International Journal of Robotics Research}, vol.~31, no.~1, pp.
  42--62, 2012.

\bibitem[Rasmussen and Williams(2006)]{Rasmussen2006}
C.~Rasmussen and C.~Williams, \emph{Gaussian Processes for Machine
  Learning}.\hskip 1em plus 0.5em minus 0.4em\relax MIT Press, 2006.

\bibitem[Shen et~al.(2005)Shen, Ng, and Seeger]{Shen2005}
Y.~Shen, A.~Ng, and M.~Seeger, ``{Fast Gaussian Process Regression using
  KD-Trees},'' \emph{Neural Information Processing Systems}, vol.~18, 2005.

\bibitem[Kim and Kim(2014)]{Kim2014}
S.~Kim and J.~Kim, ``Recursive bayesian updates for occupancy mapping and
  surface reconstruction,'' in \emph{IEEE International Conference on Robotics
  and Automation}, 2014.

\bibitem[Tresp(2000)]{VOlker2000}
V.~Tresp, ``{A Bayesian Committee Machine},'' \emph{Neural computation},
  vol.~12, pp. 2719--41, 12 2000.

\bibitem[Wang and Englot(2016)]{Wang2016}
J.~Wang and B.~Englot, ``{Fast, accurate Gaussian process occupancy maps via
  test-data octrees and nested Bayesian fusion},'' in \emph{IEEE International
  Conference on Robotics and Automation}.\hskip 1em plus 0.5em minus
  0.4em\relax IEEE, 2016, pp. 1003--1010.

\bibitem[O'Callaghan and Ramos(2011)]{OCallaghan2011}
S.~O'Callaghan and F.~Ramos, ``{Continuous Occupancy Mapping with Integral
  Kernels},'' in \emph{AAAI Conference on Artificial Intelligence}.\hskip 1em
  plus 0.5em minus 0.4em\relax AAAI, 2011.

\bibitem[Reid et~al.(2013)Reid, Ramos, and Sukkarieh]{Reid2013}
A.~Reid, F.~Ramos, and S.~Sukkarieh, ``Bayesian fusion for multi-modal aerial
  images.'' in \emph{Robotics: Science and Systems}, 2013, pp. 1--8.

\bibitem[Chen et~al.(2015)Chen, Shuai, and Chen]{Chen2015}
Y.~Chen, W.~Shuai, and X.~Chen, ``A probabilistic, variable-resolution and
  effective quadtree representation for mapping of large environments,'' in
  \emph{International Conference on Advanced Robotics}.\hskip 1em plus 0.5em
  minus 0.4em\relax IEEE, 2015, pp. 605--610.

\bibitem[S{\"{a}}rkk{\"{a}}(2011)]{Sarkka2011}
S.~S{\"{a}}rkk{\"{a}}, ``{Linear operators and stochastic partial differential
  equations in Gaussian process regression},'' \emph{Lecture Notes in Computer
  Science}, vol. 6792, no.~2, pp. 151--158, 2011.

\bibitem[Sun et~al.(2015)Sun, Vidal-Calleja, and Miro]{Sun2015}
L.~Sun, T.~Vidal-Calleja, and J.~V. Miro, ``{Bayesian fusion using
  conditionally independent submaps for high resolution 2.5D mapping},'' in
  \emph{IEEE International Conference on Robotics and Automation}, vol.
  2015-June, no. June, 2015, pp. 3394--3400.

\end{thebibliography}

\end{document}